\documentclass{ieeeaccess}
\usepackage{cite}
\usepackage{amsmath,amssymb,amsfonts}
\usepackage{algorithmic}
\usepackage{graphicx}
\usepackage{textcomp}

\usepackage{amsmath}
\usepackage{amssymb}
\usepackage{booktabs}
\usepackage{multirow}
\usepackage{enumitem}
\usepackage{array}
\usepackage{threeparttable}
\usepackage{pifont}
\usepackage{url}

\newcommand{\best}[1]{{\textbf{#1}}}
\newcommand{\second}[1]{{\underline{#1}}}
\definecolor{lightyellow}{rgb}{1,1,0.7}
\definecolor{lightred}{rgb}{1,0.75,0.75}

\makeatletter
\AtBeginDocument{\DeclareMathVersion{bold}
\SetSymbolFont{operators}{bold}{T1}{times}{b}{n}
\SetSymbolFont{NewLetters}{bold}{T1}{times}{b}{it}
\SetMathAlphabet{\mathrm}{bold}{T1}{times}{b}{n}
\SetMathAlphabet{\mathit}{bold}{T1}{times}{b}{it}
\SetMathAlphabet{\mathbf}{bold}{T1}{times}{b}{n}
\SetMathAlphabet{\mathtt}{bold}{OT1}{pcr}{b}{n}
\SetSymbolFont{symbols}{bold}{OMS}{cmsy}{b}{n}
\renewcommand\boldmath{\@nomath\boldmath\mathversion{bold}}}
\makeatother

\def\BibTeX{{\rm B\kern-.05em{\sc i\kern-.025em b}\kern-.08em
    T\kern-.1667em\lower.7ex\hbox{E}\kern-.125emX}}

\begin{document}
\vol{14}
\year{2026}
\history{Date of publication xxxx 00, 0000, date of current version xxxx 00, 0000.}
\doi{}

\title{CITRAS: Covariate-Informed Transformer for Time Series Forecasting}
\author{\uppercase{Yosuke Yamaguchi}\authorrefmark{1}, \uppercase{Issei Suemitsu}\authorrefmark{1}, and \uppercase{Wenpeng Wei}\authorrefmark{1}}

\address[1]{Research \& Development Group, Hitachi, Ltd., Tokyo 185-8601, Japan}

\tfootnote{The final version is published in IEEE Access and is available at: \url{https://doi.org/10.1109/ACCESS.2026.3695717}}

\markboth
{Y. Yamaguchi \headeretal: CITRAS: Covariate-Informed Transformer for Time Series Forecasting}
{Y. Yamaguchi \headeretal: CITRAS: Covariate-Informed Transformer for Time Series Forecasting}
\corresp{Corresponding author: Yosuke Yamaguchi (e-mail: yosuke.yamaguchi.fy@hitachi.com).}

\begin{abstract} 
In time series forecasting, covariates represent external factors that influence target variables. Some covariates are observable only in the past (observed covariates, such as recorded weather data), while others are known in advance (known covariates, such as calendar events or discount schedules). Although covariates have the potential to enhance forecasting performance, most deep learning–based forecasting models struggle to address the length discrepancy between variables caused by the future portion of known covariates and fail to leverage them flexibly. Moreover, capturing dependencies between target variables and covariates is non-trivial, as models must accurately reflect the local impact of covariates while simultaneously modeling global cross-variate dependencies. To address these challenges, we propose CITRAS, a decoder-only Transformer that flexibly integrates multiple target variables, observed covariates, and known covariates. While preserving strong autoregressive modeling capabilities, CITRAS introduces two novel mechanisms within patch-wise cross-variate attention: Key–Value (KV) Shift and Attention Score Smoothing. KV Shift seamlessly incorporates the future portion of known covariates into the forecasting process by aligning them with target variables based on their concurrent dependencies. Attention Score Smoothing refines locally accurate patch-wise cross-variate dependencies into global variate-level dependencies by smoothing the historical attention scores. Experimentally, CITRAS demonstrates strong performance across a wide range of real-world datasets in both covariate-informed and multivariate settings, showcasing its versatile ability to leverage cross-variate and cross-time dependencies for improved forecasting accuracy.

\end{abstract}


\titlepgskip=-21pt

\maketitle

\section{Introduction}
\label{sec:intro}
\begin{figure*}[t]
    \centering
    \includegraphics[trim=10 10 10 10, clip, width=0.9\textwidth]{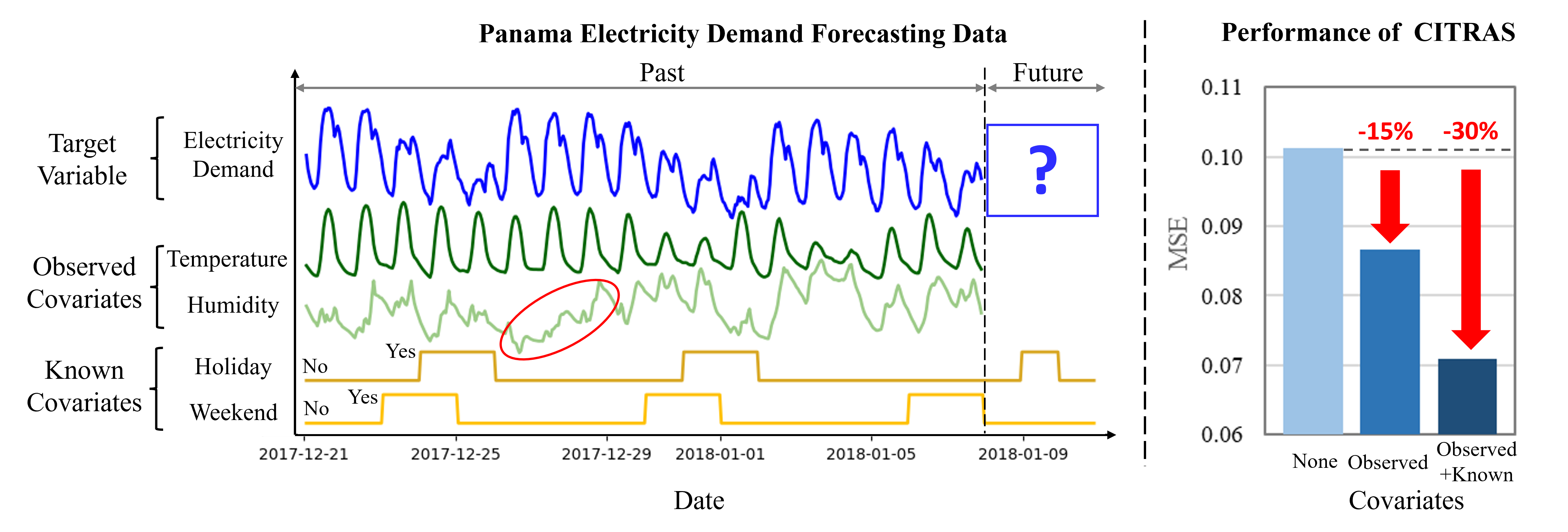}
    \caption{Left: Hourly electricity demand data for three weeks in Panama \cite{madrid2021elec}. ``Holiday'' and ``Weekend'' indicators demonstrate a negative correlation with demand, providing important contextual information in the future forecasting horizon. ``Humidity'' displays a weak correlation but introduces temporal disturbances (red circle), which may distract the forecasting process. Right:~Performance improvement of CITRAS through the utilization of observed covariates and known covariates.}
    \label{fig:teaser}
\end{figure*}

Time series forecasting is a cornerstone in diverse fields such as energy \cite{weron2014electricity}, retail \cite{bose2017demand}, and finance \cite{capistran2010inflation}, where accurate predictions can drive strategic decisions and enhance operational efficiency. In practical scenarios, forecasters have access not only to the target variables they aim to predict but also to covariates that represent external factors influencing the target variables. For instance, the ``Electricity Demand'' data shown in Figure \ref{fig:teaser} (Left) demonstrates a strong negative correlation with the ``Holiday'' indicator, which is accessible throughout the future forecasting horizon. This highlights the critical importance of appropriately incorporating covariates into the forecasting process.

Driven by these practical demands, this paper addresses time series forecasting that involves two types of covariates. 
The first is the \textit{observed covariate}, whose values are available only in the historical period up to the prediction time, such as recorded weather information.
The second is the \textit{known covariate}, whose values are available from the past through the entire forecasting horizon. 
Known covariates may represent predetermined quantities (e.g., calendar events), estimated information (e.g., weather forecasts), or controllable variables (e.g., planned discounts in retail). 

This practical problem setting presents two significant challenges. First, the forecasting model must accommodate heterogeneous variables flexibly. The availability and number of covariates can vary depending on the scenario, and known covariates exhibit a length discrepancy with other variables. In addition, many real-world applications require forecasting multiple target variables simultaneously, i.e., \textit{multivariate} forecasting.
Second, the model must capture the dependencies between variables from both fine-grained and coarse-grained perspectives. The fine-grained perspective enables the model to precisely capture the local impact of covariates, such as sudden drops in electricity demand on holidays.
However, a model that relies solely on a fine-grained perspective may fail to capture global dependencies between variables, as it can overlook covariates representing infrequent events or can be distracted by temporal disturbances. Therefore, it is crucial for the model to capture dependencies between variables at both fine-grained and coarse-grained levels to ensure robust and accurate forecasting.

In recent years, deep forecasting models—particularly Transformers~\cite{vaswani2017attention}—have achieved remarkable progress owing to their strong ability to capture temporal dependencies. However, most existing approaches still fall short in handling heterogeneous covariates. Many are tailored exclusively to multivariate forecasting~\cite{zhang2022crossformer,wang2024card}, others depend solely on observed covariates~\cite{liu2025timerxl}, and some are limited to known covariates~\cite{tayal2024exotst}. These limitations suggest that the extension of Transformers to flexibly incorporate both observed and known covariates, while still preserving their strong temporal modeling capabilities, remains an open direction for further exploration.

Existing approaches to capture cross-variate dependencies in Transformers can be broadly categorized into two types: variate-level and patch-level. iTransformer \cite{liu2023itransformer} is a representative variate-level approach that captures inter-variate correlations by embedding each sequence into a single variate token and applying attention across multiple variate tokens. On the other hand, patch-level approaches, represented by Crossformer~\cite{zhang2022crossformer}, apply cross-variate attention between locally semantic tokens obtained through patching. While variate-level approaches excel at capturing global dependencies robustly to noisy interactions, patch-level approaches are better at capturing fine-grained information. Unfortunately, an architecture that enjoys the advantages of both of these approaches in cross-variate modeling is yet to be developed.

To overcome these challenges, we propose CITRAS: \textbf{C}ovariate-\textbf{I}nformed \textbf{Tra}n\textbf{s}former for time series forecasting that flexibly leverages multiple target variables, observed covariates, and known covariates. This model is a decoder-only patch-based Transformer that has a cross-time attention module and a cross-variate attention module separately. To effectively model the cross-variate dependencies at both fine-grained and coarse-grained perspectives, we introduce Attention Score Smoothing in the cross-variate attention module, which refines locally accurate patch-wise dependencies into global variate-level dependencies. Furthermore, we introduce the Key-Value~(KV) Shift mechanism, which associates the key of the known covariate with the value from one patch step ahead, thereby seamlessly integrating future information from known covariates into the prediction process along with current information from observed covariates. These innovations expand the covariate understanding capabilities of the canonical decoder-only Transformer, fully preserving its strong autoregressive properties.

In summary, our contributions are as follows: \begin{itemize} 
    \item To effectively exploit the external impacts represented by covariates, we extend the decoder-only Transformer to flexibly accommodate observed covariates and known covariates while fully maintaining its original strong autoregressive capability.
    \item Our model, CITRAS, introduces two novel mechanisms into the cross-variate attention module. Attention Score Smoothing captures locally accurate patch-wise dependencies and refines them into global variate-level dependencies, while KV Shift seamlessly integrates future information from known covariates based on the obtained dependencies.
    \item Experimentally, CITRAS demonstrates strong forecasting performance in both covariate-informed and multivariate settings, ranking first more frequently than other state-of-the-art models by effectively leveraging cross-variate and cross-time dependencies.
\end{itemize}

\section{Related Work}\label{sec:relatedwork}
\subsection{Transformer-based Time Series Forecasting}
Given the significant success of Transformers in the fields of natural language processing \cite{devlin2018bert} and computer vision \cite{dosovitskiy2021vit}, numerous studies have attempted to leverage their capabilities for time series forecasting. These works can be roughly categorized into two approaches: channel-dependent (CD) and channel-independent (CI). The CD approach assumes that the future values of a variable are determined by its past values as well as the values of other variables. In contrast, the CI approach posits that a specific variable depends only on its past values, omitting explicit interactions between variables.

Early applications of Transformers in time series forecasting can be viewed as CD approaches, where multiple variables at each time step are embedded into temporal tokens \cite{zhou2022fedformer,liu2021pyraformer,dong2023simmtm}. To address the computational costs and intricate dependencies arising from long time series data, Informer~\cite{zhou2021informer} designs a ProbSparse self-attention mechanism with a distilling operation to efficiently focus on important features. Autoformer~\cite{wu2021autoformer} incorporates decomposition and an auto-correlation mechanism to uncover reliable dependencies from complex temporal patterns.
However, point-wise temporal tokens have limited local semantics, making it difficult to capture intricate cross-time and cross-variate dependencies.

In response to these challenges, PatchTST~\cite{nie2022patchtst} adopts a CI approach and introduces patching, which handles time series data by segmenting it into fixed-length segments. Patching reduces computational costs and enhances the local semantics in token representation, making it a prevalent technique in subsequent methods \cite{zhang2024upme,zhang2024elastst,wang2024sthd}. Additionally, the CI approach is also adopted by many recent models as it avoids the risk of unnecessary noisy interactions between channels.
For example, DLinear~\cite{zeng2023dlinear} decomposes each time series into trend and seasonal components and applies simple linear projections independently to each channel. FITS~\cite{xu2024fits} transforms each univariate time series into the frequency domain via the Fourier transform, learning spectral representations separately for each variable before inverse transformation for forecasting. Moreover, recent large models \cite{ansari2024chronos,das2023decoder,goswami2024moment} and LLM-based forecasters \cite{finzi2023llmtime,zhou2023one,jin2024timellm} also adopt the CI approach. 

However, the CI approach is sometimes regarded as an oversimplification \cite{abdelmalak2025fact}, especially considering that fluctuations in time series data are often influenced by external factors. A representative CD approach is iTransformer~\cite{liu2023itransformer}, which embeds each channel into a single variate token and applies cross-variate attention over them. TimeXer~\cite{wang2024timexer} and Leddam~\cite{yu2024leddam} also capture cross-variate dependencies at the variate level, with an additional cross-time attention to capture intra-variate temporal dependencies. TQNet~\cite{lin2025tqnet} captures cross-variate dependencies both from variate- and dataset-level perspectives by introducing learnable query vectors for each channel used throughout the entire dataset.
Meanwhile, Crossformer~\cite{zhang2022crossformer} applies cross-variate attention to patch-level representations. Any-variate attention techniques \cite{woo2024moirai,liu2025timerxl,liu2025moiraimoe} also apply attention to patch-level representations obtained by flattening all variables into one sequence. While the patch-level approach captures more fine-grained dependencies between variables, its local receptive field lacks global context awareness. Our method, CITRAS, first captures fine-grained dependencies at the patch level, and Attention Score Smoothing gradually refines them into global, variate-level dependencies. 

\subsection{Covariate-Informed Time Series Forecasting} 
To leverage the rich contextual information provided by covariates, numerous approaches have been proposed. Statistical models such as ARIMAX~\cite{williams2001multivariate} or VARX~\cite{nicholson2017varx} extend their original frameworks by assuming linear relationships between target variables and covariates. Recently, many deep learning models focus on covariate-informed settings:
N-BEATSx~\cite{olivares2023neural} enhances the original N-BEATS~\cite{oreshkin2019nbeats} by introducing an additional MLP-based residual stack for covariates.
TiDE~\cite{das2023tide} introduces an MLP-based architecture that utilizes future information of known covariates in the final projection to leverage their direct effects on future target variables. 
ExoTST~\cite{tayal2024exotst} treats past and future information of known covariates as separate modalities and fuses them via cross-time attention to enrich autoregressive forecasting with projected covariate information.
Timer-XL~\cite{liu2025timerxl} introduces a masking scheme, TimeAttention, in a decoder-only Transformer to model covariate influence within any-variate attention. 
DeformTime~\cite{shu2025deformtime} captures both inter-variable and intra-variable relationships by employing deformable attention to dynamically adjust receptive fields across variables and time.
More recently, Sonnet~\cite{shu2025sonnet} proposes a Spectral Operator Neural Network that models dependencies between variables in the frequency domain by applying learnable wavelet transforms and an attention mechanism that measures inter-variable relationships based on spectral coherence rather than time-domain similarity.

\begin{table}[t]
\centering
\begin{threeparttable}
\caption{Comparison of Base Architecture and Supported Variable Types for Covariate-Informed Forecasting}
\label{tab:model_spec}
\begin{tabular}{l c c c c c}
\toprule
\multirow{2}{*}{Models} & \multicolumn{1}{c}{Base} & \multicolumn{1}{c}{Uni-} & \multicolumn{1}{c}{Multi-} &
\multicolumn{1}{c}{Obs.} &
\multicolumn{1}{c}{Kno.} \\
 & \multicolumn{1}{c}{arch.} & \multicolumn{1}{c}{variate} & \multicolumn{1}{c}{variate} & \multicolumn{1}{c}{cov.} & \multicolumn{1}{c}{cov.} \\
\midrule
CITRAS (ours)     & TR(Dec) & \ding{51} & \ding{51} & \ding{51} & \ding{51} \\
\midrule
TFT~\cite{lim2021temporal}     & TR+RNN & \ding{51} & \ding{51} & \ding{51} & \ding{51} \\
TSMixer-Ext~\cite{chen2023tsmixer}     & MLP & \ding{51} & \ding{51} & \ding{51} & \ding{51} \\
TimeXer~\cite{wang2024timexer}   & TR & \ding{51} & \ding{51}  & \ding{51}  & \ding{71} \\
N-BEATSx~\cite{olivares2023neural}     & MLP & \ding{51} & \ding{55}  & \ding{55}  & \ding{51} \\
TiDE~\cite{das2023tide}     & MLP & \ding{51} & \ding{55}  & \ding{55}  & \ding{51} \\
ExoTST~\cite{tayal2024exotst}     & TR & \ding{51} & \ding{55}  & \ding{55}  & \ding{51} \\
Timer-XL~\cite{liu2025timerxl}  & TR(Dec) & \ding{51} & \ding{51}  & \ding{51}  & \ding{55} \\
DeformTime~\cite{shu2025deformtime}   & TR+RNN & \ding{51} & \ding{51}  & \ding{51}  & \ding{55} \\
Sonnet~\cite{shu2025sonnet}   & TR & \ding{51} & \ding{55}  & \ding{51}  & \ding{55} \\
\bottomrule
\end{tabular}
\begin{tablenotes}[flushleft]\footnotesize
\item The symbol \ding{71} in TimeXer indicates its numerical evaluation with known covariates is not originally reported.
\item Arch.: Architecture; TR: Transformer; TR(Dec): Transformer (Decoder-only); Obs. cov.: Observed covariate; Kno. cov.: Known covariate.
\end{tablenotes}
\end{threeparttable}
\end{table}

However, these models fall short in flexibly handling multiple targets, observed covariates, and known covariates, as shown in Table~\ref{tab:model_spec}.
Indeed, Timer-XL~\cite{liu2025timerxl}, DeformTime~\cite{shu2025deformtime}, and Sonnet~\cite{shu2025sonnet} rely solely on observed covariates, whereas N-BEATSx~\cite{olivares2023neural}, TiDE~\cite{das2023tide}, and ExoTST~\cite{tayal2024exotst} can handle only known covariates as they always require future covariate information.
Although TimeXer~\cite{wang2024timexer} is architecturally capable of handling both types of covariates, its numerical evaluations focus exclusively on settings with observed covariates. Our experiments indicate that its variate-level dependency modeling design fails to accurately reflect the localized impact of known covariates. More importantly, this design aggregates information from the entire temporal span into each variable token, making it incompatible with the autoregressive causal structure inherent to decoder-only Transformers such as CITRAS.

A few exceptions, such as the Temporal Fusion Transformer (TFT)~\cite{lim2021temporal} and TSMixer-Ext~\cite{chen2023tsmixer}, can handle all variable types. TFT combines recurrent encoders with an interpretable attention mechanism to integrate both types of covariates within a unified framework. It employs gating layers and variable-selection networks to dynamically filter relevant features and utilizes a multi-head attention decoder to capture long-range temporal dependencies.
TSMixer-Ext~\cite{chen2023tsmixer} consists of MLP-based mixer layers along both temporal and variate dimensions, embedding observed and known covariates into a shared representation early in the model.
However, these models adopt complex architectures to integrate heterogeneous variables, which are incompatible with decoder-only Transformers. Consequently, as highlighted in Table~\ref{tab:model_spec}, there is still no established method for extending decoder-only Transformers—known for their strong autoregressive capability—to flexibly incorporate both observed and known covariates.
CITRAS addresses this gap by seamlessly integrating both types of covariates into a decoder-only Transformer via its KV Shift mechanism, fully preserving its strong autoregressive performance.

\section{CITRAS}\label{sec:method}
\subsection{Problem Setting}
Let $\mathbf{X}_{1:T}^{tgt,:}=\{ \mathbf{X}_{1:T}^{tgt,1},\mathbf{X}_{1:T}^{tgt,2},..., \mathbf{X}_{1:T}^{tgt,C_{tgt}} \} \in \mathbb{R}^{T \times C_{tgt}}$ be a multivariate target time series of length $T$ with $C_{tgt}$ variables. With optional usage of observed covariates $\mathbf{X}_{1:T}^{obs,:} \in \mathbb{R}^{T \times C_{obs}}$ and known covariates $\mathbf{X}_{1:T+S}^{knw,:} \in \mathbb{R}^{(T+S) \times C_{knw}}$, the goal of the forecasting model $\mathcal{F}_\theta$ is to predict the future $S$ time steps of the target time series, which can be formulated as follows:
\begin{equation}
    \widehat{\mathbf{X}}_{T+1:T+S}^{tgt,:} = \mathcal{F}_\theta\left( \mathbf{X}_{1:T}^{tgt,:},\mathbf{X}_{1:T}^{obs,:}, \mathbf{X}_{1:T+S}^{knw,:} \right)
\end{equation}

\subsection{Architecture}
\begin{figure*}[t]
    \centering
    \includegraphics[trim=0 0 0 0, clip, width=0.85\textwidth]{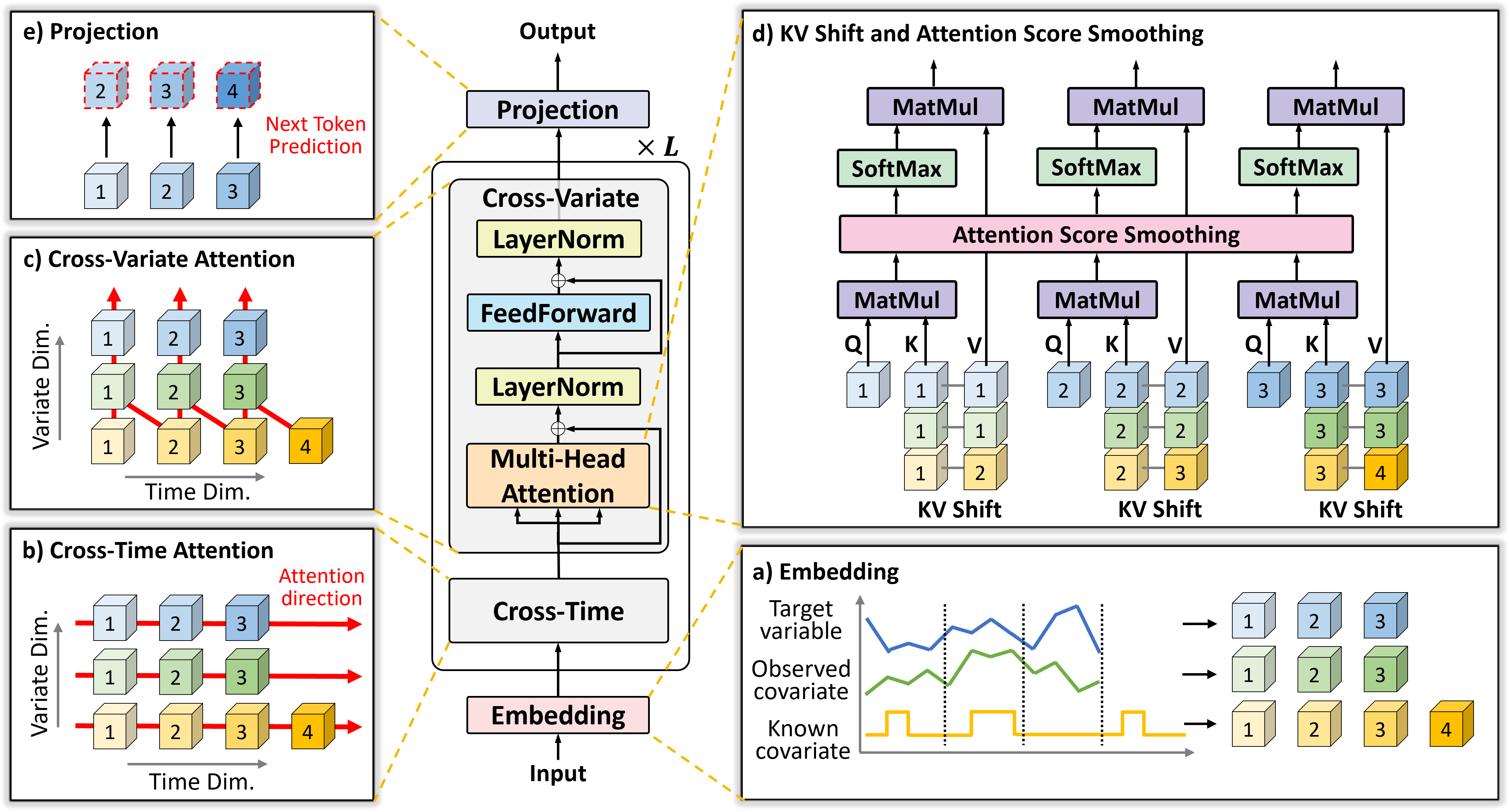}
    \caption{Overall structure of CITRAS. (a) The embedding module applies patching to each variable, yielding temporal token embeddings. (b) The cross-time attention module captures intra-variate cross-time dependencies. (c) The cross-variate attention module captures cross-variate dependencies. (d) KV Shift associates the key of the known covariate with the value from one patch step ahead. Attention Score Smoothing calculates the Exponential Moving Average (EMA) of patch-wise attention scores. (e) The projection layer projects each target token embedding to the values of the next patch.}
    \label{fig:arc}
\end{figure*}

As shown in Figure~\ref{fig:arc}, CITRAS is a patch-based, decoder-only Transformer with $L$ layers of cross-time attention modules and cross-variate attention modules. Our core mechanisms, KV Shift and Attention Score Smoothing, reside in the cross-variate attention module.

\subsubsection{Embedding}
First, all sequences are segmented into non-overlapping patches of length $P$, and each patch is embedded into a $D$-dimensional token. For simplicity, we assume that both $T$ and $S$ are divisible by $P$. Taking a target variable $c$ as an example, this can be formalized as:
\begin{equation}
\begin{aligned}
    \{\mathbf{s}_1^{tgt,c},\mathbf{s}_2^{tgt,c},...,\mathbf{s}_{N_{tgt}}^{tgt,c} \} = \operatorname{Patchify} \left(\mathbf{X}_{1:T}^{tgt,c} \right)\\
    \mathbf{H}_i^{tgt,c} = \operatorname{Embed}\left(\mathbf{s}_i^{tgt,c}\right), ~~i=1,...,N_{tgt}
\end{aligned}    
\end{equation}
where $N_{tgt} = \frac{T}{P}$ is the number of patches and $\operatorname{Embed}:\mathbb{R}^P \to \mathbb{R}^D$ is a shared linear projector across all variables. We denote the token embeddings of a target variable at all patch steps as $ \mathbf{H}_{:}^{tgt,c}=~\{ \mathbf{H}_i^{tgt,c} \}_{i=1}^{N_{tgt}}~\in~\mathbb{R}^{N_{tgt} \times D}$. Similarly, the token embeddings of an observed covariate are denoted as $\mathbf{H}_{:}^{obs,c} \in \mathbb{R}^{N_{obs} \times D}$, and those of a known covariate are denoted as $\mathbf{H}_{:}^{knw,c} \in \mathbb{R}^{N_{knw} \times D}$, all obtained using the shared parameters with the target variables. Here, $N_{obs} = N_{tgt} = \frac{T}{P}$ and $N_{knw} =\frac{(T+S)}{P}$.

\subsubsection{Cross-Time Attention}
In the cross-time attention module, we apply multi-head attention with causal masking to all variables to capture their intra-variate cross-time dependencies. Following Timer-XL~\cite{liu2025timerxl}, we adopt Rotary Position Embedding (RoPE) \cite{jianlin2024roformer} to capture temporal order. Taking a target variable $c$ as an example, and dropping the layer index for brevity, this can be formalized as:

\begin{equation}
\begin{aligned}
\widetilde{\mathbf{H}}_{:}^{tgt,c} &= \operatorname{LN}\left( \mathbf{H}_{:}^{tgt,c} + \operatorname{MHA}\left(\mathbf{H}_{:}^{tgt,c},\mathbf{H}_{:}^{tgt,c},\mathbf{H}_{:}^{tgt,c}\right)\right)\\
\mathbf{H}_{:}^{tgt,c} &= \operatorname{LN}\left( \widetilde{\mathbf{H}}_{:}^{tgt,c} + \operatorname{FFN}\left(\widetilde{\mathbf{H}}_{:}^{tgt,c}\right)\right)\\
\end{aligned}
\end{equation}
where $\operatorname{LN}$ denotes layer normalization \cite{ba2016layernormalization}, $\operatorname{MHA}\left(\mathbf{Q},\mathbf{K},\mathbf{V}\right)$ denotes the multi-head attention layer where $\mathbf{Q}$, $\mathbf{K}$, and $\mathbf{V}$ serve as queries, keys, and values, and $\operatorname{FFN}$ denotes a feed-forward network. 

Again, token embeddings of observed covariates $\mathbf{H}_{:}^{obs,c}$ and those of known covariates $\mathbf{H}_{:}^{knw,c}$ are similarly processed with the shared parameters.

\subsubsection{Cross-Variate Attention --- KV Shift} Attention is recognized as an effective approach for capturing cross-variate dependencies \cite{liu2023itransformer}. To represent the dependencies of target variables while avoiding unnecessary interactions among covariates, we adopt multi-head attention, where queries originate from the target variables, whereas keys and values are derived from all available variables. However, applying attention across the variables at each patch step, as adopted by Crossformer~\cite{zhang2022crossformer}, fails to leverage pivotal future patches of known covariates, as they do not have corresponding target patches.

To address this issue, we introduce a KV Shift mechanism in the multi-head attention layer of the cross-variate attention module, which associates the key of the known covariate with the value from one patch step ahead. Specifically, the key and value at patch step $i$ can be formalized as:
\begin{equation}
\begin{aligned}
     \text{Key:}~~~\mathbf{H}_i^{k,:} &= \left[\mathbf{H}_i^{tgt,:},\mathbf{H}_i^{obs,:},\mathbf{H}_i^{knw,:} \right]\\
    \text{Value:}~~~\mathbf{H}_i^{v,:} &= \left[\mathbf{H}_i^{tgt,:},\mathbf{H}_i^{obs,:},\mathbf{H}_{i+1}^{knw,:} \right]\\
\end{aligned}
\end{equation}
where $\left[ \cdot,\cdot\right]$ denotes the concatenation along the variate dimension and $\mathbf{H}_i^{k,:}, \mathbf{H}_i^{v,:} \in \mathbb{R}^{\left(C_{tgt}+C_{obs}+C_{knw}\right) \times D}$. After this, cross-variate dependencies can be captured by the standard multi-head attention layer:
\begin{equation}
\begin{aligned}
    \widetilde{\mathbf{H}}_i^{tgt,:} &= \operatorname{LN}\left(\mathbf{H}_i^{tgt,:} + \operatorname{MHA}\left(\mathbf{H}_i^{tgt,:},\mathbf{H}_i^{k,:}, \mathbf{H}_i^{v,:}\right)\right) \\
    \mathbf{H}_i^{tgt,:} &= \operatorname{LN}\left( \widetilde{\mathbf{H}}_i^{tgt,:} + \operatorname{FFN}\left(\widetilde{\mathbf{H}}_i^{tgt,:}\right)\right)
\end{aligned}
\end{equation}
for $i=1,...,N_{tgt}$. KV Shift maintains step alignment in the dot product calculation between queries and keys, enabling precise capture of concurrent dependencies.
As a result, the model first identifies how strongly each known covariate influences the target variable at the current step.
It then allows the target token to incorporate the next-step future known covariate (value) based on the strength of these obtained dependencies.
As the target token gradually transforms into the prediction for the next step, this facilitates a natural flow of information for exploiting future information from known covariates.
Note that KV Shift is applied exclusively to attention from target queries to known covariates, and never to target–target or target–observed-covariate interactions, thereby preventing unrealistic access to future information or temporal leakage.

\subsubsection{Cross-Variate Attention --- Attention Score Smoothing}
The attention score of $\operatorname{MHA}\left(\mathbf{Q},\mathbf{K},\mathbf{V}\right) $ is calculated as:
\begin{equation}
\widetilde{\mathbf{A}} = \left(\mathbf{Q}\mathbf{W}_q \right) \left(\mathbf{K}\mathbf{W}_k\right)^\top
\end{equation}
where $\mathbf{W}_q, \mathbf{W}_k \in \mathbb{R}^{D \times d_k}$ and $d_k$ is the dimension of the query, key, and value.
In the cross-variate attention module, this attention score is calculated for each patch step separately, capturing fine-grained dependencies between variables at that step. However, time series data often exhibit local disturbances, and sparse variables representing infrequent events may maintain constant values within a patch. In such scenarios, local patch-wise attention fails to represent global variate-level dependencies.

To address this issue, we introduce Attention Score Smoothing in the multi-head attention layer of the cross-variate attention module, which transforms locally accurate patch-wise dependencies into global variate-level dependencies. Specifically, the attention score at each step is smoothed based on the attention scores up to that step in each head. We employ an Exponential Moving Average (EMA) as a smoothing method, as it enables adaptation to shifting correlations between variables \cite{kang2025vardrop} by assigning exponentially decreasing weights over time. Denoting original attention score at patch step $i$ as $\widetilde{\mathbf{A}}_i$, the smoothed attention score $\mathbf{A}_i$ can be calculated as:
\begin{equation}
\begin{aligned}
\mathbf{A}_i &= \alpha \widetilde{\mathbf{A}}_i + \left(1-\alpha\right)\mathbf{A}_{i-1},~~i=2,...,N_{tgt} \\
\end{aligned}
\end{equation}
where $\mathbf{A}_1=\widetilde{\mathbf{A}}_1$ and $\alpha$ is a smoothing factor, which we consider as a shared hyperparameter across all cross-variate attention modules and heads. Its design is inspired by temporal smoothing and regularization principles in time series analysis, where aggregating information over time reduces variance in noisy estimates. By applying an exponential moving average to patch-wise attention scores, the model stabilizes cross-variate dependency estimation while remaining adaptive to gradual changes.

It is important to note that the objective and mechanism of Attention Score Smoothing are completely different from Exponential Smoothing Attention in ETSformer \cite{woo2022etsformer}. Exponential Smoothing Attention aims for cross-time dependency modeling. It replaces the original token-similarity-based attention score with exponentially decreasing weights over time. In contrast, Attention Score Smoothing is for cross-variate dependency modeling. We first utilize the original attention to obtain cross-variate dependencies based on token similarity at each patch step, and then smooth these patch-wise dependencies over time to obtain global variate-level dependencies.

\subsubsection{Projection}
Following the next-token prediction approach common in decoder-only Transformers, each token embedding of the target variables is used to predict the values of the next patch. For $i=1,...,N_{tgt}$ and $c=1,...,C_{tgt}$, this can be formalized as:
\begin{equation}
     \widehat{\mathbf{X}}_{iP+1:(i+1)P}^{tgt,c} = \operatorname{Project}\left( \mathbf{H}_i^{tgt,c} \right)
\end{equation}
where $\operatorname{Project}: \mathbb{R}^D \to \mathbb{R}^P$ is a shared linear projector across all steps and target variables. In the training phase, we calculate the squared loss using all of these outputs. In the testing phase, the outputs from the last patches are used for forecasting. When the forecasting horizon $S$ exceeds the patch length $P$, the output target values are integrated into subsequent inputs for recursive forecasting.  

\section{Experiments}\label{sec:experiment}
To verify the effectiveness and versatility of CITRAS, we extensively evaluated it in two settings, covariate-informed forecasting and well-established multivariate forecasting.

\subsection{Covariate-Informed Forecasting}\label{sec:covariate-forecasting}
\subsubsection{Dataset} 
We use seven real-world datasets for covariate-informed forecasting, with detailed information provided in Table~\ref{tab:dataset-detail}. 
They include target variable(s), observed covariates, and known covariates, thereby providing a realistic representation of practical forecasting scenarios. 

\textbf{EPF} datasets~\cite{lago2021forecasting} consist of five subsets from different day-ahead electricity markets, each spanning six years: 
\textbf{EPF-NP} documents the Nord Pool electricity market from 2013-01-01 to 2018-12-24, containing hourly electricity prices as a target variable, with corresponding grid load and wind power forecasts as known covariates. 
\textbf{EPF-PJM} records the Pennsylvania-New Jersey-Maryland market from 2013-01-01 to 2018-12-24, containing zonal electricity prices in the Commonwealth Edison (COMED) as a target variable, with corresponding system load and COMED load forecasts as known covariates. 
\textbf{EPF-BE} captures Belgium’s electricity market from 2011-01-09 to 2016-12-31, containing hourly electricity prices as a target variable, with corresponding load forecasts in Belgium and generation forecasts in France as known covariates. 
\textbf{EPF-FR} documents the electricity market in France from 2012-01-09 to 2017-12-31, containing hourly prices as a target variable, with corresponding load and generation forecasts as known covariates. 
\textbf{EPF-DE} records the German electricity market from 2012-01-09 to 2017-12-31, containing hourly prices as a target variable, with zonal load forecasts in the TSO Amprion zone and wind and solar generation forecasts as known covariates. 

\textbf{EDF}~\cite{madrid2021elec} is an hourly time series dataset spanning over five years from 2015-01-03 to 2020-06-27, where the target variable is the national electricity demand in Panama. Observed covariates include four meteorological indicators (air temperature, specific humidity, total precipitable liquid water, and wind speed) measured in three cities (Tocumen, Santiago, David) in Panama, resulting in a total of 12 variables. In addition, we adopt three binary calendar features (holiday, school day, weekend) as known covariates. 

\textbf{BS}~\cite{fan2013bike_sharing} is an hourly time series dataset spanning two years from 2011-01-01 to 2012-12-31, where the target variables are three types of rental counts (casual, registered, and total count) in the bike sharing system in Washington, D.C. Observed covariates include five meteorological indicators (weather situation, temperature, feeling temperature, humidity, and windspeed). In addition, we adopt three calendar features (holiday, weekday, working day) as known covariates.

Following previous work~\cite{olivares2023neural}, we set the input length as 168 and forecasting length as 24.

\begin{table*}[h]
\caption{Dataset descriptions.}
\begin{threeparttable}
\centering
{\fontsize{9pt}{11pt}\selectfont
\begin{tabular}{ccccccc}
\toprule[1.2pt]
Task & Dataset & \#Target & \#Observed & \#Known & Sampling Frequency & Dataset Size \\ \toprule[1.2pt]
\multirow{3}{*}{\shortstack{Covariate-informed\\Forecasting}}& EPFs & 1 & 0 & 2 & 1 Hour & (36500, 5219, 10460) \\ 
\cmidrule(l){2-7}
& EDF & 1 & 12 & 3 & 1 Hour & (33442, 4783, 9586) \\ \cmidrule(l){2-7}
& BS & 3 & 5 & 3 & 1 Hour & (11974, 1716, 3452) \\ \midrule
\multirow{7}{*}{\shortstack{Multivariate\\Forecasting}} & ETTh & 7 & 0 & 0 & 1 Hour &(8545, 2881, 2881) \\ \cmidrule(l){2-7}
 & ETTm & 7 & 0 & 0 & 15 Minutes & (34465, 11521, 11521) \\ \cmidrule(l){2-7}
& ECL & 321 & 0 & 0 & 1 Hour & (18317, 2633, 5261) \\ \cmidrule(l){2-7}
& Weather & 21 & 0 & 0 & 10 Minutes & (36792, 5271, 10540) \\
\cmidrule(l){2-7}
& Traffic & 862 & 0 & 0 & 1 Hour & (12185, 1757, 3509) \\
\cmidrule(l){2-7}
& PEMS04 & 307 & 0 & 0 & 5 Minutes & (10100, 3400, 3399) \\
\cmidrule(l){2-7}
& PEMS08 & 170 & 0 & 0 & 5 Minutes & (10618, 3573, 3572) \\

\bottomrule[1.2pt]
\end{tabular}}
\begin{tablenotes}[para,flushleft]
\#Target, \#Observed, and \#Known refer to the number of target variables, observed covariates, and known covariates, respectively. The dataset size is presented as (Train, Validation, Test).
\end{tablenotes}
\label{tab:dataset-detail}
\end{threeparttable}
\end{table*}

\subsubsection{Baselines}\label{sec:covariate-baselines}
\begin{table}[t]
\centering
\begin{threeparttable}
\caption{Comparison of Supported Variable Types for Baseline Models, Including Our Extensions.}
\label{tab:baseline_spec}
\begin{tabular}{l c c c c}
\toprule
\multirow{2}{*}{Models} & \multicolumn{1}{c}{Uni-} & \multicolumn{1}{c}{Multi-} &
\multicolumn{1}{c}{Observed} &
\multicolumn{1}{c}{Known} \\
 &  \multicolumn{1}{c}{variate} & \multicolumn{1}{c}{variate} & \multicolumn{1}{c}{covariate} & \multicolumn{1}{c}{covariate} \\
\midrule
TFT~\cite{lim2021temporal}      & \multirow{3}{*}{\ding{51}} & \multirow{3}{*}{\ding{51}} & \multirow{3}{*}{\ding{51}} & \multirow{3}{*}{\ding{51}} \\
TSMixer-Ext~\cite{chen2023tsmixer}     & & & & \\
TimeXer~\cite{wang2024timexer}   & & & & \\
\midrule
TiDE~\cite{das2023tide}     & \ding{51} & \ding{55}  & \ding{55}  & \ding{51} \\
\midrule
Timer-XL~\cite{liu2025timerxl}  & \multirow{1}{*}{\ding{51}} & \multirow{1}{*}{\ding{51}} & \multirow{1}{*}{\ding{51}} & \multirow{1}{*}{\ding{71}} \\
\midrule
iTransformer~\cite{liu2023itransformer} & \multirow{2}{*}{\ding{51}} & \multirow{2}{*}{\ding{51}} & \multirow{2}{*}{\ding{70}} & \multirow{2}{*}{\ding{71}} \\
Leddam~\cite{yu2024leddam} & & & & \\
\midrule
CARD~\cite{wang2024card} & \multirow{4}{*}{\ding{51}} & \multirow{4}{*}{\ding{51}} & \multirow{4}{*}{\ding{70}} & \multirow{4}{*}{\ding{55}} \\
ModernTCN~\cite{donghao2024moderntcn} & & & & \\
TimesNet~\cite{wu2023timesnet} & & & & \\
Crossformer~\cite{zhang2022crossformer} & & & & \\
\midrule
FITS~\cite{xu2024fits} & \multirow{3}{*}{\ding{51}} & \multirow{3}{*}{\ding{55}} & \multirow{3}{*}{\ding{55}} & \multirow{3}{*}{\ding{55}} \\
DLinear~\cite{zeng2023dlinear} & & & & \\
PatchTST~\cite{nie2022patchtst} & & & & \\
\bottomrule
\end{tabular}
\begin{tablenotes}[flushleft]\footnotesize
\item The symbol \ding{70} indicates that the model treats the target and observed covariates jointly without explicit distinction. The symbol \ding{71} indicates that the model was extended by us to support known covariates in our experiments.
\end{tablenotes}
\end{threeparttable}
\end{table}

We include 14 baseline models, and Table~\ref{tab:baseline_spec} summarizes the variable types supported by each model.
TFT~\cite{lim2021temporal}, TSMixer-Ext~\cite{chen2023tsmixer}, and TimeXer~\cite{wang2024timexer} originally support multivariate, observed covariates, and known covariates, while TiDE~\cite{das2023tide} supports univariate with known covariates. These models are used in their original forms.
Timer-XL~\cite{liu2025timerxl} originally supports only observed covariates, so we extend it to use known covariates. Specifically, we modify the masking scheme in TimeAttention so that target token embeddings can attend to one-step ahead future token embeddings of known covariates.
Since iTransformer~\cite{liu2023itransformer}, Leddam~\cite{yu2024leddam}, CARD~\cite{wang2024card}, ModernTCN~\cite{donghao2024moderntcn}, TimesNet~\cite{wu2023timesnet}, and Crossformer~\cite{zhang2022crossformer} are originally multivariate models, they inherently treat the target and observed covariates jointly without explicit distinction.
Among them, we additionally extend iTransformer and Leddam to incorporate known covariates by introducing a variate embedding layer to accommodate their longer sequence lengths. The details of the extension of Timer-XL, iTransformer, and Leddam are provided in the Appendix (Section~\ref{sec:extension}).
In contrast, FITS~\cite{xu2024fits}, DLinear~\cite{zeng2023dlinear}, and PatchTST~\cite{nie2022patchtst} are univariate models and they are used in their original forms.

Future information from known covariates is expected to directly affect the forecasting accuracy. 
For a fair comparison with baselines that cannot utilize known covariates, we evaluate CITRAS both with and without known covariates (i.e., without using KV Shift). 
In the setting without known covariates, the future portion of known covariates is omitted and treated similarly to observed covariates.
We reproduce all baseline implementations based on the TimesNet~\cite{wu2023timesnet} repository or their official repositories.

\subsubsection{Implementation Details}
All experiments were implemented in PyTorch~\cite{paszke2019pytorch}.
We utilize a single NVIDIA V100 32GB GPU for covariate-informed forecasting tasks.
We apply Series Stationarization~\cite{liu2022non} to the embedding and projection layers of CITRAS to mitigate nonstationarity.
We adopt Adam~\cite{kingma2014adam} with an initial learning rate of $10^{-4}$ and L2 loss for model optimization. The training process is conducted for up to 10 epochs with an early stopping mechanism, which terminates training if the validation performance does not improve for 3 consecutive epochs.
We set the number of layers in our proposed model $L \in \{1, 2, 3, 4\}$, the number of heads in the multi-head attention layers as 8, embedding dimension $D \in \{128, 256, 512, 1024\}$, and the smoothing factor for Attention Score Smoothing $\alpha \in \{0.1, 0.2, 0.4, 0.8\}$. 
All hyperparameters are selected based on the validation performance using a grid search approach. 
Hyperparameters of the baseline models are tuned within the ranges reported in their respective original papers following the same procedure.
For all models, the patch length is uniformly fixed to 24.

We ensure that the drop last option is set to False in the test data loader to avoid any unfair impact, as pointed out by~\cite{qiu2024tfb}. 
All evaluations in covariate-informed forecasting tasks are conducted with three random seeds and the average performance is reported. 
The standard deviation is provided in the Appendix (Section~\ref{sec:covariate-std}).

\subsubsection{Results}\label{sec:covariate-result}
Table~\ref{tab:covariate-result} presents the results of covariate-informed forecasting with and without the use of known covariates. In the former setting, CITRAS outperforms TFT, TSMixer-Ext, and TiDE, all of which employ complex architectures to handle future information from known covariates. This superior performance can be attributed to the simple yet effective design of CITRAS, which seamlessly integrates the future information from covariates into the decoder-only Transformer through KV Shift, thereby preserving its inherent expressive power. Among recent models that adhere to the canonical Transformer architecture, TimeXer captures cross-variate dependencies at the variate level, while Timer-XL does so at the patch level. The relative superiority of these contrasting methods is dataset-dependent. In contrast, CITRAS leverages the advantages of both approaches through Attention Score Smoothing, consistently demonstrating strong performance across datasets. Moreover, the superiority of CITRAS remains evident even in conventional settings without known covariates. This is because multivariate baselines treat target variables and covariates equally and suffer from unnecessary interactions among covariates. Also, the univariate baselines cannot account for target fluctuations caused by observed covariates, leading to poor modeling of cross-time dependencies.
In contrast, CITRAS applies attention only from target variables to covariates, thereby effectively utilizing observed covariates without causing unnecessary interactions.

\begin{figure}[h]
    \includegraphics[trim=5 5 5 5, clip, width=1.0\linewidth]{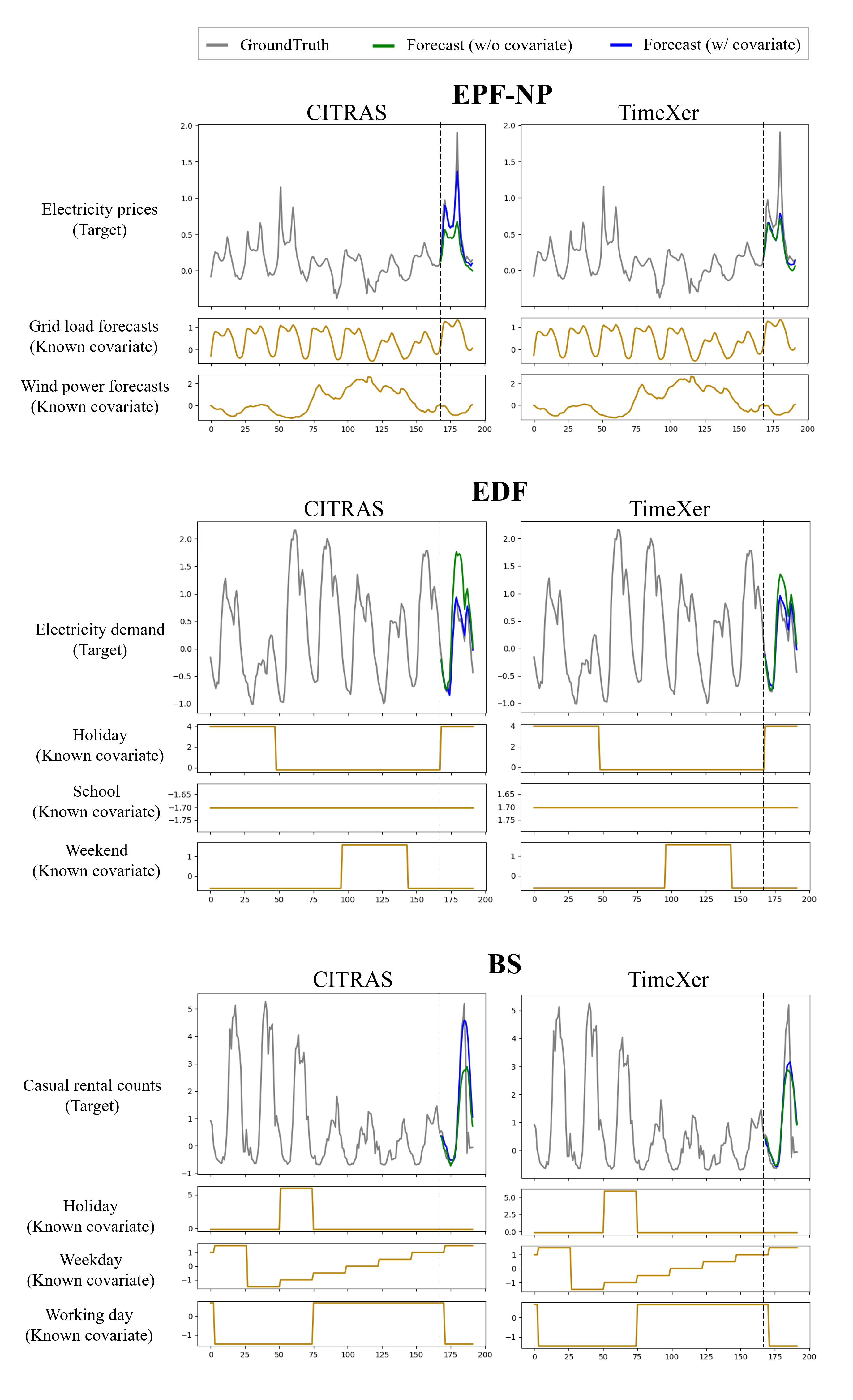}
    \caption{Forecasting examples of CITRAS and TimeXer without (shown in green) and with (shown in blue) the use of covariates. The target variable is displayed in the top box, while the known covariates are shown in the boxes below.}
    \label{fig:vis_main}
\end{figure}

Figure~\ref{fig:vis_main} presents the visualization results for the EPF-NP, EDF, and BS datasets. Forecasts without any covariates are shown in green, while those with all covariates are shown in blue.
In the EPF-NP result, CITRAS accurately predicts the sudden rise in target electricity prices by taking into account the high grid load and low power generation forecasts during the forecasting horizon.
In the EDF result, it adjusts the electricity demand downward by recognizing that the forecasting horizon falls on a holiday.
In the BS result, it effectively captures the negative impact of the binary working day covariate and accurately forecasts the increase in rental counts on the day following a working day.
These results indicate that CITRAS does not simply incorporate covariates that resemble the target shape, but rather captures complex relationships from a global perspective and accurately reflects the local impact of future known covariates.
In contrast, TimeXer leverages cross-variate dependencies only at the variate level, and thus struggles to adjust the forecasts locally accurately.
These comparisons highlight the strength of CITRAS that effectively utilizes covariates at multiple granularities.
\begin{table*}[ht]
\begin{threeparttable}
\caption{Results of the covariate-informed forecasting with and without known covariates. Average MSE and MAE across three random seeds are reported.}
\label{tab:covariate-result}
\setlength{\tabcolsep}{3.5pt}
\centering
{\fontsize{9pt}{10.5pt}\selectfont
\begin{tabular}
{p{10pt}|c|cccccccccccccccc} 

\toprule[1.2pt]
\multicolumn{2}{c|}{{Model}}   &
\multicolumn{2}{c}{{\textbf{CITRAS}}} &
\multicolumn{2}{c}{{TFT}} &
\multicolumn{2}{c}{{TSMixer-Ext}} &
\multicolumn{2}{c}{{TimeXer}} &
\multicolumn{2}{c}{{TiDE}} &
\multicolumn{2}{c}{{Timer-XL}} &
\multicolumn{2}{c}{{iTrans.}} &
\multicolumn{2}{c}{{Leddam}}\\

\cmidrule(lr){3-4}\cmidrule(lr){5-6}\cmidrule(lr){7-8}\cmidrule(lr){9-10}
\cmidrule(lr){11-12}\cmidrule(lr){13-14}\cmidrule(lr){15-16}\cmidrule(lr){17-18}
\multicolumn{2}{c|}{{Metric}} &
{MSE} & {MAE} &
{MSE} & {MAE} &
{MSE} & {MAE} &
{MSE} & {MAE} &
{MSE} & {MAE} &
{MSE} & {MAE} &
{MSE} & {MAE} &
{MSE} & {MAE}\\ 

\toprule
\multirow{7}{*}{\vspace{-28pt} \rotatebox{90}{\scalebox{1.4}{\hspace{15pt}w/ known}}}
& {EPF--NP}
& \best{{0.172}} & \best{{0.215}}
& {0.193} & {0.236}
& {0.187} & {0.251}
& {0.180} & {0.228}
& {0.298} & {0.311}
& \second{{0.175}} & \second{{0.218}}
& {0.211} & {0.254}
& {0.254} & {0.285} \\

& {EPF-PJM}
& \best{{0.063}} & \best{{0.153}}
& {0.084} & {0.177}
& {0.078} & {0.172}
& {0.081} & {0.176}
& {0.106} & {0.214}
& \second{{0.065}} & \second{{0.157}}
& {0.077} & {0.170}
& {0.090} & {0.196} \\

& {EPF--BE}
& {0.363} & {0.248}
& {0.386} & \second{{0.245}}
& \best{{0.343}} & {0.249}
& {0.368} & \best{{0.232}}
& {0.449} & {0.302}
& {0.422} & {0.274}
& \second{{0.345}} & {0.247}
& {0.372} & {0.255} \\

& {EPF--FR}
& \best{{0.360}} & \best{{0.176}}
& {0.373} & \second{{0.188}}
& {0.426} & {0.216}
& \second{{0.366}} & {0.191}
& {0.411} & {0.253}
& {0.429} & {0.199}
& {0.384} & {0.207}
& {0.418} & {0.222} \\

& {EPF--DE}
& \best{{0.219}} & \best{{0.293}}
& {0.270} & {0.331}
& {0.252} & {0.324}
& {0.293} & {0.324}
& {0.522} & {0.467}
& \second{{0.229}} & \second{{0.301}}
& {0.263} & {0.329}
& {0.283} & {0.337} \\

& {EDF}
& \best{{0.071}} & \best{{0.194}}
& {0.090} & {0.224}
& {0.080} & {0.212}
& \second{{0.071}} & {0.198}
& {0.135} & {0.261}
& {0.081} & {0.208}
& {0.072} & \second{{0.197}}
& {0.080} & {0.207} \\

& {BS}
& \best{{0.282}} & \best{{0.314}}
& {0.375} & {0.377}
& {0.344} & {0.358}
& {0.340} & {0.356}
& {0.490} & {0.447}
& \second{{0.286}} & \second{{0.317}}
& {0.350} & {0.355}
& {0.383} & {0.388} \\

\toprule[1.2pt]
\multicolumn{2}{c|}{{Model}}   &
\multicolumn{2}{c}{{\textbf{CITRAS}}} &
\multicolumn{2}{c}{{CARD}} &
\multicolumn{2}{c}{{ModernTCN}} &
\multicolumn{2}{c}{{TimesNet}} &
\multicolumn{2}{c}{{Cross.}} &
\multicolumn{2}{c}{{FITS}} &
\multicolumn{2}{c}{{DLinear}} &
\multicolumn{2}{c}{{PatchTST}}\\

\cmidrule(lr){3-4}\cmidrule(lr){5-6}\cmidrule(lr){7-8}\cmidrule(lr){9-10}\cmidrule(lr){11-12}\cmidrule(lr){13-14}\cmidrule(lr){15-16}\cmidrule(lr){17-18}
\multicolumn{2}{c|}{{Metric}} & {MSE} & {MAE} & {MSE}    & {MAE}   & {MSE}    & {MAE}    & {MSE}    & {MAE} & {MSE}    & {MAE} & {MSE}    & {MAE} & {MSE}    & {MAE} & {MSE}    & {MAE}\\ 

\toprule
\multirow{7}{*}{\vspace{-28pt} \rotatebox{90}{\scalebox{1.4}{\hspace{13pt}w/o known}}} & {EPF--NP} & \best{{0.227}} & \best{{0.260}} & {0.266} & {0.293} & \second{{0.237}} & \second{{0.274}} & {0.247} & {0.283} & {0.248} & {0.288} & {0.304} & {0.316} & {0.309} & {0.321} & {0.264} & {0.287} \\

& {EPF-PJM} & \best{{0.090}} & \best{{0.182}} & {0.114} & {0.218} & \second{{0.100}} & \second{{0.195}} & {0.101} & {0.201} & {0.104} & {0.198} & {0.109} & {0.216} & {0.108} & {0.214} & {0.108} & {0.213} \\

& {EPF--BE} & \second{{0.405}} & {0.276} & {0.425} & {0.287} & \best{{0.393}} & \second{{0.267}} & {0.416} & {0.280} & {0.414} & {0.283} & {0.460} & {0.308} & {0.463} & {0.314} & {0.408} & \best{{0.260}} \\

& {EPF--FR} & {0.407} & {0.219} & {0.425} & {0.249} & {0.410} & {0.231} & \second{{0.406}} & {0.221} & {0.433} & \best{{0.213}} & \best{{0.403}} & {0.255} & {0.429} & {0.260} & {0.412} & \second{{0.215}} \\

& {EPF--DE} & \best{{0.420}} & \best{{0.404}} & {0.470} & {0.434} & \second{{0.445}} & \second{{0.416}} & {0.484} & {0.435} & {0.525} & {0.423} & {0.533} & {0.473} & {0.523} & {0.465} & {0.463} & {0.431} \\

& {EDF} & \best{{0.087}} & \best{{0.206}} & {0.089} & \second{{0.212}} & \second{{0.088}} & {0.216} & {0.108} & {0.238} & {0.112} & {0.236} & {0.134} & {0.258} & {0.133} & {0.259} & {0.103} & {0.232} \\

& {BS} & \best{{0.289}} & \best{{0.321}} & {0.414} & {0.416} & {0.377} & {0.401} & {0.394} & {0.382} & \second{{0.339}} & \second{{0.361}} & {0.484} & {0.439} & {0.469} & {0.429} & {0.349} & {0.377} \\

\bottomrule[1.2pt]

\end{tabular}}
\begin{tablenotes}[para,flushleft]
    The input length and forecasting length are set to 168 and 24, respectively. The best result is represented in bold, followed by underline.
\end{tablenotes}
\end{threeparttable}

\end{table*}

\subsection{Multivariate Forecasting}
\label{sec:multi}
For a comprehensive comparison, we further evaluate our model on well-established multivariate forecasting benchmarks. 

\subsubsection{Dataset}
We use nine real-world datasets for multivariate forecasting without using any covariates, with detailed information provided in Table~\ref{tab:dataset-detail}.

\begin{itemize}
    \item \textbf{ETT}~\cite{zhou2021informer}: Contains 7 monitoring factors in electricity transformers from July 2016 to July 2018. \textbf{ETTh1} and \textbf{ETTh2} are hourly subsets, and \textbf{ETTm1} and \textbf{ETTm2} are subsets recorded every 15 minutes.

    \item \textbf{Weather}~\cite{wu2021autoformer}: Includes 21 meteorological factors recorded every 10 minutes from the Weather Station of the Max Planck Biogeochemistry Institute in 2020.

    \item \textbf{ECL}~\cite{wu2021autoformer}: Comprises hourly electricity consumption data from 321 clients.

    \item \textbf{Traffic}~\cite{wu2021autoformer}: Records hourly road occupancy rates measured by 862 sensors on San Francisco Bay area freeways from January 2015 to December 2016.

    \item \textbf{PEMS04}~\cite{liu2022scinet}: Contains public traffic network data recorded at 307 locations in California every 5 minutes from 2018-01-01 to 2018-02-28.

    \item \textbf{PEMS08}~\cite{liu2022scinet}: Contains data at 170 locations from 2016-07-01 to 2016-08-31.

\end{itemize}

We use all the variables in these datasets as target variables.
To reflect practical scenarios, we adopt the rolling forecasting approach~\cite{liu2024autotimes}, where a model trained with predetermined input and forecasting lengths is evaluated with a longer forecasting length by iteratively integrating its output into the subsequent input.
Following previous works~\cite{liu2024autotimes,liu2025timerxl}, we set the input length to 672, the training forecasting length to 96, and the testing forecasting length to \{96, 192, 336, 720\}.

\subsubsection{Implementation Details}
We utilize a single NVIDIA A100 80GB GPU for the ECL, Traffic, and PEMS datasets and a single NVIDIA V100 32GB GPU for the other datasets.
The patch length is uniformly set to 96.
All evaluations in multivariate forecasting tasks are conducted with a fixed random seed and a single run, following the standard practice in previous works~\cite{wang2024timexer, liu2025timerxl}. 
Other settings are the same as those in covariate-informed forecasting.

\subsubsection{Results}
Table~\ref{tab:multi-full-results} presents the results of multivariate forecasting.
Compared to recent models specifically designed for multivariate settings, CITRAS achieves superior or competitive performance across most datasets.
In particular, it performs strongly on the ECL, Traffic, PEMS04, and PEMS08 datasets, which involve a large number of target variables and require both accurate and scalable modeling of complex inter-variable relationships.
Overall, CITRAS ranks first more frequently than any other model, highlighting the effectiveness of its design and its robustness across diverse forecasting scenarios.
\begin{table*}[htbp]
\begin{threeparttable}
\caption{Full results of the multivariate forecasting}
\label{tab:multi-full-results}
\renewcommand{\multirowsetup}{\centering}
\setlength{\tabcolsep}{1pt}
\centering
\begin{tabular*}{\textwidth}{@{\extracolsep{\fill}}
    c c *{9}{cc}}
\toprule[1.2pt]  
\multicolumn{2}{c}{{Models}}  & \multicolumn{2}{c}{{\textbf{CITRAS}}} & \multicolumn{2}{c}{{TimeXer}} & \multicolumn{2}{c}{{Timer-XL}} & \multicolumn{2}{c}{{iTrans.}} & \multicolumn{2}{c}{{Leddam}} & \multicolumn{2}{c}{{CARD}} & \multicolumn{2}{c}{\scalebox{1}{ModernTCN}} & \multicolumn{2}{c}{{FITS}} & \multicolumn{2}{c}{{DLinear}}\\ 
\cmidrule(lr){1-2}\cmidrule(lr){3-4}\cmidrule(lr){5-6}\cmidrule(lr){7-8}\cmidrule(lr){9-10}\cmidrule(lr){11-12}\cmidrule(lr){13-14}\cmidrule(lr){15-16}\cmidrule(lr){17-18}\cmidrule(lr){19-20}
\multicolumn{2}{c}{{Metric}}   & {MSE} & {MAE} & {MSE}    & {MAE}   & {MSE}    & {MAE}    & {MSE}    & {MAE} & {MSE}    & {MAE} & {MSE}    & {MAE}  & {MSE} & {MAE}      & {MSE}    & {MAE} & {MSE}    & {MAE}\\ 
\toprule[1.2pt]

\multirow{5}{*}{{\rotatebox{90}{ETTh1}}}
& {96} & \best{{0.355}} & \best{{0.395}} & {0.391} & {0.425} & \second{{0.364}} & \second{{0.397}} & {0.387} & {0.418} & {0.371} & {0.406} & {0.373} & {0.403} & {0.372} & {0.400} & {0.378} & {0.401} & {0.369} & {0.400} \\ 
& {192} & \best{{0.386}} & \best{{0.416}} & {0.420} & {0.443} & {0.405} & {0.424} & {0.416} & {0.437} & \second{{0.397}} & {0.422} & {0.402} & {0.422} & {0.405} & {0.421} & {0.410} & \second{{0.420}} & {0.405} & {0.422} \\
& {336} & \best{{0.403}} & \best{{0.429}} & {0.441} & {0.459} & {0.427} & {0.439} & {0.434} & {0.450} & \second{{0.417}} & {0.435} & {0.424} & {0.436} & {0.428} & {0.437} & {0.431} & \second{{0.433}} & {0.435} & {0.445} \\
& {720} & \best{{0.412}} & \best{{0.445}} & {0.460} & {0.483} & \second{{0.439}} & {0.459} & {0.447} & {0.473} & {0.441} & {0.463} & {0.459} & {0.472} & {0.446} & {0.468} & {0.441} & \second{{0.451}} & {0.493} & {0.508} 

\\ \cmidrule(lr){2-20} & {Ave} & \best{{0.389}} & \best{{0.421}} & {0.428} & {0.453} & {0.409} & {0.430} & {0.421} & {0.445} & \second{{0.406}} & {0.431} & {0.414} & {0.433} & {0.413} & {0.432} & {0.415} & \second{{0.426}} & {0.426} & {0.444} \\  \midrule

\multirow{5}{*}{{\rotatebox{90}{ETTh2}}}
& {96} & {0.293} & {0.358} & {0.281} & {0.350} & {0.301} & {0.354} & {0.307} & {0.365} & \second{{0.275}} & {0.345} & {0.282} & {0.345} & \best{{0.271}} & \best{{0.340}} & {0.277} & \second{{0.342}} & {0.286} & {0.354} \\
& {192} & {0.351} & {0.395} & {0.339} & {0.391} & {0.360} & {0.396} & {0.376} & {0.406} & \second{{0.330}} & {0.381} & {0.338} & {0.381} & \best{{0.327}} & \second{{0.378}} & {0.334} & \best{{0.377}} & {0.344} & {0.393} \\
& {336} & {0.372} & {0.416} & {0.371} & {0.419} & {0.382} & {0.419} & {0.416} & {0.435} & \second{{0.355}} & {0.402} & {0.361} & {0.401} & \best{{0.351}} & \second{{0.399}} & {0.358} & \best{{0.398}} & {0.369} & {0.416} \\
&{720} & {0.419} & {0.453} & {0.422} & {0.459} & {0.443} & {0.468} & {0.432} & {0.456} & {0.394} & {0.434} & \best{{0.380}} & \best{{0.423}} & {0.396} & {0.433} & \second{{0.387}} & \second{{0.426}} & {0.408} & {0.454} \\
\cmidrule(lr){2-20}
& {Ave} & {0.359} & {0.405} & {0.353} & {0.405} & {0.372} & {0.409} & {0.383} & {0.415} & \second{{0.339}} & {0.391} & {0.340} & \second{{0.388}} & \best{{0.336}} & \second{{0.388}} & \second{{0.339}} & \best{{0.386}} & {0.352} & {0.404} \\ \midrule

\multirow{5}{*}{{\rotatebox{90}{ETTm1}}}
& {96} & \best{{0.282}} & \best{{0.340}} & {0.300} & {0.358} & \second{{0.297}} & \second{{0.347}} & {0.312} & {0.367} & {0.303} & {0.351} & {0.305} & {0.351} & {0.311} & {0.360} & {0.308} & {0.351} & {0.307} & {0.350} \\
& {192} & \best{{0.329}} & {0.369} & {0.342} & {0.381} & {0.344} & {0.376} & {0.351} & {0.390} & {0.339} & {0.372} & \second{{0.336}} & {0.370} & {0.347} & {0.380} & {0.338} & \best{{0.367}} & {0.337} & \second{{0.368}} \\ 
& {336} & \best{{0.365}} & {0.391} & {0.383} & {0.402} & {0.382} & {0.399} & {0.389} & {0.412} & {0.370} & {0.392} & \best{{0.365}} & {0.387} & {0.381} & {0.400} & {0.367} & \best{{0.384}} & {0.366} & \second{{0.386}} \\
& {720} & {0.423} & {0.425} & {0.450} & {0.436} & {0.450} & {0.437} & {0.458} & {0.449} & {0.423} & {0.422} & \best{{0.415}} & \second{{0.416}} & {0.438} & {0.433} & \second{{0.418}} & \best{{0.413}} & \second{{0.418}} & {0.418} \\
\cmidrule(lr){2-20} & {Ave} & \best{{0.350}} & {0.382} & {0.368} & {0.394} & {0.368} & {0.390}  & {0.377} & {0.404} & {0.358} & {0.384} & \second{{0.355}} & {0.381} & {0.369} & {0.393} & {0.358} & \best{{0.379}} & {0.357} & \second{{0.380}}  \\ \midrule

\multirow{5}{*}{{\rotatebox{90}{ETTm2}}}
& {96} & {0.174} & {0.260} & {0.176} & {0.263} & {0.180} & {0.263} & {0.183} & {0.272} & \second{{0.163}} & \second{{0.253}} & \best{{0.162}} & \best{{0.252}} & {0.178} & {0.269} & {0.164} & {0.255} & {0.166} & {0.259} \\
& {192} & {0.231} & {0.299}& {0.235} & {0.301} & {0.244} & {0.306}  & {0.241} & {0.310} & \second{{0.218}} & \second{{0.292}} & \best{{0.214}} & \best{{0.287}} & {0.235} & {0.307} & {0.219} & {0.293} & {0.222} & {0.300} \\
& {336} & {0.280} & {0.333}& {0.287} & {0.335} & {0.299} & {0.344}  & {0.296} & {0.346} & \second{{0.271}} & \second{{0.328}} & \best{{0.263}} & \best{{0.319}} & {0.288} & {0.340} & {0.272} & \second{{0.328}} & {0.278} & {0.339} \\
& {720} & {0.365} & {0.389} & {0.368} & {0.386} & {0.377} & {0.397} & {0.393} & {0.409} & {0.363} & {0.385} & \best{{0.351}} & \best{{0.375}} & {0.366} & {0.388} & \second{{0.360}} & \second{{0.382}} & {0.379} & {0.405} \\
\cmidrule(lr){2-20}
& {Ave} & {0.262} & {0.320} & {0.266} & {0.321} & {0.275} & {0.327} & {0.278} & {0.334} & \second{{0.254}} & {0.315} & \best{{0.247}} & \best{{0.308}} & {0.267} & {0.326} & \second{{0.254}} & \second{{0.314}} & {0.262} & {0.326} \\ \midrule

\multirow{5}{*}{{\rotatebox{90}{ECL}}}
&  {96} & \second{{0.128}} & \second{{0.222}}& {0.131} & {0.231} & \best{{0.127}} & \best{{0.219}}  & {0.133} & {0.229} & {0.130} & {0.224} & {0.129} & {0.225} & {0.161} & {0.270} & {0.142} & {0.244} & {0.138} & {0.238} \\
& {192} & {0.149} & {0.242} & {0.149} & {0.248} & \best{{0.145}} & \best{{0.236}} & {0.158} & {0.258} & {0.148} & \second{{0.241}} & \second{{0.147}} & {0.242} & {0.172} & {0.281} & {0.156} & {0.256} & {0.152} & {0.251} \\
& {336} & \second{{0.164}} & {0.259} & {0.168} & {0.268} & \best{{0.159}} & \best{{0.252}} & {0.168} & {0.262} & \second{{0.164}} & \second{{0.257}} & \second{{0.164}} & {0.259} & {0.183} & {0.292} & {0.173} & {0.272} & {0.167} & {0.268} \\
& {720} & \second{{0.199}} & \second{{0.288}} & {0.214} & {0.309} & \best{{0.187}} & \best{{0.277}} & {0.205} & {0.294} & {0.202} & {0.291} & {0.202} & {0.292} & {0.216} & {0.321} & {0.213} & {0.304} & {0.203} & {0.302} \\
\cmidrule(lr){2-20} & {Ave} & \second{{0.160}} & \second{{0.253}} & {0.166} & {0.264} & \best{{0.155}} & \best{{0.246}} & {0.164} & {0.258} & {0.161} & \second{{0.253}} & \second{{0.160}} & {0.255} & {0.183} & {0.291} & {0.171} & {0.269} & {0.165} & {0.265} \\ \midrule

\multirow{5}{*}{{\rotatebox{90}{Weather}}}
& {96} & {0.152} & {0.205} & {0.150} & {0.203} & {0.157} & {0.205} & {0.174} & {0.225} & {0.149} & \second{{0.201}} & \second{{0.147}} & {0.204} & {0.153} & {0.210} & \best{{0.145}} & \best{{0.197}} & {0.169} & {0.229} 
\\ & {192} & {0.200} & {0.249} & {0.192} & \second{{0.242}} & {0.206} & {0.250} & {0.227} & {0.268} & {0.194} & {0.243} & \second{{0.191}} & {0.244} & {0.199} & {0.250} & \best{{0.186}} & \best{{0.237}} & {0.211} & {0.268} \\ & {336} & {0.253} & {0.291} & \second{{0.240}} & \second{{0.280}} & {0.259} & {0.291} & {0.290} & {0.309} & {0.242} & {0.283} & \best{{0.238}} & {0.282} & {0.251} & {0.290} & {0.241} & \best{{0.277}} & {0.258} & {0.306} \\ & {720} & {0.325} & {0.342} &\best{ {0.309}} & \best{{0.329}} & {0.337} & {0.344} & {0.374} & {0.360} & {0.314} & {0.335} & \best{{0.309}} & \second{{0.334}} & {0.323} & {0.341} & {0.342} & {0.347} & {0.320} & {0.362} \\ 
\cmidrule(lr){2-20} & {Ave} & {0.233} & {0.272} & \second{{0.223}} & \best{{0.263}} & {0.240} & {0.273} & {0.266} & {0.291} & {0.225} & \second{{0.265}} & \best{{0.221}} & {0.266} & {0.232} & {0.273} & {0.229} & \second{{0.265}} & {0.239} & {0.291} \\ \midrule

\multirow{5}{*}{{\rotatebox{90}{Traffic}}}
& {96} & \best{{0.338}} & \best{{0.235}} & {0.362} & {0.264} & \second{{0.340}} & \second{{0.238}} & {0.353} & {0.259} & {0.356} & {0.256} & {0.373} & {0.266} & {0.432} & {0.320} & {0.390} & {0.275} & {0.399} & {0.285} \\ & {192} & \second{{0.365}} & \best{{0.247}} & {0.381} & {0.275} & \best{{0.360}} & \best{{0.247}} & {0.373} & {0.267} & {0.376} & {0.264} & {0.386} & {0.272} & {0.434} & {0.321} & {0.402} & {0.279} & {0.409} & {0.290} \\ & {336} & {0.392} & \second{{0.258}} & {0.402} & {0.288} & \best{{0.377}} & \best{{0.256}} & \second{{0.386}} & {0.275} & {0.390} & {0.271} & {0.401} & {0.279} & {0.441} & {0.323} & {0.415} & {0.285} & {0.422} & {0.297} \\ & {720} & \second{{0.419}} & \best{{0.276}} & {0.457} & {0.324} & \best{{0.418}} & \second{{0.279}} & {0.425} & {0.296} & {0.423} & {0.290} & {0.441} & {0.303} & {0.476} & {0.344} & {0.453} & {0.306} & {0.461} & {0.319} \\ 
\cmidrule(lr){2-20} & {Ave} & \second{{0.379}} & \best{{0.254}} & {0.401} & {0.288} & \best{{0.374}} & \second{{0.255}} & {0.384} & {0.274} & {0.386} & {0.270} & {0.400} & {0.280} & {0.446} & {0.327} & {0.415} & {0.286} & {0.423} & {0.298} \\ \midrule

\multirow{5}{*}{{\rotatebox{90}{PEMS04}}}
& {96} & \second{{0.105}} & \best{{0.201}} & {0.123} & {0.238} & \best{{0.104}} & \second{{0.207}} & {0.111} & {0.216} & \second{{0.105}} & {0.209} & {0.141} & {0.248} & {0.125} & {0.245} & {0.212} & {0.306} & {0.196} & {0.296}
\\ & {192} & {0.118} & \best{{0.212}} & {0.136} & {0.249} & \second{{0.117}} & \second{{0.218}} & {0.123} & {0.225} & \best{{0.115}} & \second{{0.218}} & {0.162} & {0.264} & {0.133} & {0.251} & {0.229} & {0.317} & {0.214} & {0.310}
\\ & {336} & {0.130} & \best{{0.221}} & {0.150} & {0.262} & \second{{0.129}} & \second{{0.228}} & {0.135} & {0.236} & \best{{0.127}} & \second{{0.228}} & {0.187} & {0.285} & {0.142} & {0.260} & {0.252} & {0.334} & {0.236} & {0.328} 
\\ & {720} & \best{{0.151}} & \best{{0.244}} & {0.180} & {0.295} & \second{{0.155}} & \second{{0.255}} & {0.168} & {0.269} & {0.158} & {0.258} & {0.254} & {0.342} & {0.163} & {0.289} & {0.354} & {0.404} & {0.330} & {0.398} \\
\cmidrule(lr){2-20} & {Ave} & \best{{0.126}} & \best{{0.220}} & {0.147} & {0.261} & \best{{0.126}} & \second{{0.227}} & {0.135} & {0.237} & \best{{0.126}} & {0.228} & {0.186} & {0.285} & {0.141} & {0.261} & {0.262} & {0.340} & {0.244} & {0.333} \\ \midrule

\multirow{5}{*}{{\rotatebox{90}{PEMS08}}}
& {96} & \best{{0.140}} & \best{{0.192}} & {0.185} & {0.242} & \second{{0.163}} & \second{{0.207}} & {0.190} & {0.225} & {0.174} & {0.218} & {0.239} & {0.277} & {0.261} & {0.286} & {0.362} & {0.338} & {0.325} & {0.335} \\ 
& {192} & \best{{0.223}} & \best{{0.207}} & {0.252} & {0.261} & \second{{0.245}} & \second{{0.224}} & {0.266} & {0.240} & {0.250} & {0.233} & {0.315} & {0.302} & {0.321} & {0.305} & {0.422} & {0.357} & {0.381} & {0.358} \\
& {336} & \best{{0.282}} & \best{{0.220}} & {0.296} & {0.282} & \second{{0.283}} & \second{{0.237}} & {0.315} & {0.255} & {0.303} & {0.246} & {0.363} & {0.326} & {0.359} & {0.323} & {0.460} & {0.376} & {0.418} & {0.378} \\
& {720} & \second{{0.300}} & \best{{0.246}} & {0.346} & {0.332} & \best{{0.299}} & \second{{0.267}} & {0.349} & {0.291} & {0.333} & {0.280} & {0.403} & {0.377} & {0.406} & {0.381} & {0.526} & {0.441} & {0.506} & {0.449} \\
\cmidrule(lr){2-20} & {Ave} & \best{{0.236}} & \best{{0.216}} & {0.270} & {0.279} & \second{{0.247}} & \second{{0.234}} & {0.280} & {0.253} & {0.265} & {0.244} & {0.330} & {0.321} & {0.337} & {0.324} & {0.442} & {0.378} & {0.408} & {0.380} \\ \midrule

\multicolumn{2}{c|}{{$1^{\text{st}}$}count} & \best{{16}} & \best{{20}} & {1} & {2} & \second{{12}} & {7} & {0} & {0} & {3} & {0} & {11} & {6} & {4} & {1} & {2} & \second{{10}} & {0} & {0}\\ 

\bottomrule[1.2pt]  
\end{tabular*}
\begin{tablenotes}[para,flushleft]
    Full results of the multivariate forecasting, where the input length is set to 672, and the forecasting length is set to \{96, 192, 336, 720\}. We adopt the rolling forecasting approach ~\cite{liu2024autotimes}, where one model is used for four forecasting lengths. Results are cited from Timer-XL \cite{liu2025timerxl} if available; otherwise reproduced. 
\end{tablenotes}
\end{threeparttable}
\end{table*}

\subsection{Model Analysis}
\subsubsection{Ablation Study}\label{sec:ablation}
CITRAS introduces KV Shift and Attention Score Smoothing (ASS) on top of a decoder-only Transformer. 
To validate the effectiveness of these design choices, we conduct detailed ablation studies, as summarized in Table~\ref{tab:abl}.
In the KV Shift ablation (``w/o KV Shift''), the target token in cross-variate attention is restricted to attend only to known covariates at the same patch step, preventing access to future known covariate information. As a result, the model fails to leverage future covariates in covariate-informed forecasting, leading to a significant performance drop across datasets.
On top of this ``w/o KV Shift'' variant, we further evaluate a simpler alternative (``w/ Late fusion''), which fuses future token embeddings of known covariates with target token embeddings using an additional learnable fusion layer immediately before the final projection layer (see Appendix, Section~\ref{sec:extension} for details). Although this late-fusion strategy allows future known covariates to be included, it leads to inferior performance on datasets such as EPF-NP and BS compared to the full model. This result indicates that naive fusion is insufficient, as it bypasses the dependency modeling performed within cross-variate attention. In contrast, KV Shift integrates future known covariates directly into the attention mechanism without introducing additional parameters, enabling more principled and consistent exploitation of future information.
In the ASS ablation (``w/o ASS''), the model is restricted to capturing cross-variate dependencies only from a local perspective, resulting in inferior performance in both covariate-informed and multivariate settings.
A more in-depth analysis of the sensitivity to the smoothing factor is provided in Section~\ref{sec:ass_param}.
For the decoder-only ablation, we alternatively adopt an encoder-only architecture (i.e., removing causal masking in cross-time attention and applying a projection layer to the flattened features of all tokens for prediction). However, this design cannot effectively model causality in forecasting and lacks token-wise supervision, thus failing to fully leverage the inherent autoregressive capability of the Transformer in most cases.
Collectively, these results highlight the strength of CITRAS, which seamlessly integrates heterogeneous covariates into a decoder-only Transformer and effectively models both cross-variate and cross-time dependencies.

\begin{table}[th] 
\caption{MSE results of the ablation study. All ablated variants are evaluated using a fixed random seed. Values in parentheses indicate that the corresponding mechanism is not applicable to datasets without known covariates, and thus the base CITRAS results are duplicated.}
\setlength{\tabcolsep}{2pt}
\centering
{\fontsize{9pt}{10.5pt}\selectfont

\begin{tabular}{l|ccccc}
\toprule[1.2pt]
\multicolumn{1}{c}{{Design}} & {EPF-NP}  & {EDF} & {BS} &  {ETTh1} & {ECL} \\
\midrule

{CITRAS} & \textbf{{0.172}} & {0.071} & \textbf{{0.282}} & \textbf{{0.389}} & \textbf{{0.160}} \\

{~(w/o) KV Shift} & {0.226} & {0.086} & {0.289} & {(0.389)} & {(0.160)} \\

{~(w/ ) Late Fusion} & {0.195} & {0.071} & {0.327} & {(0.389)} & {(0.160)} \\

{~(w/o) ASS} & {0.175} & {0.074} & {0.299} & {0.401} & {0.165} \\

{~(w/o) Decoder-only} & {0.188} & \textbf{{0.068}} & {0.366} & {0.428} & {0.162} \\

\bottomrule[1.2pt]
\end{tabular}}
\label{tab:abl}
\end{table}

\subsubsection{Attention Score Smoothing}\label{sec:ass_param}
\begin{figure}[h]
    \centering
        \includegraphics[trim=8 8 8 8, clip, width=1.0\linewidth]{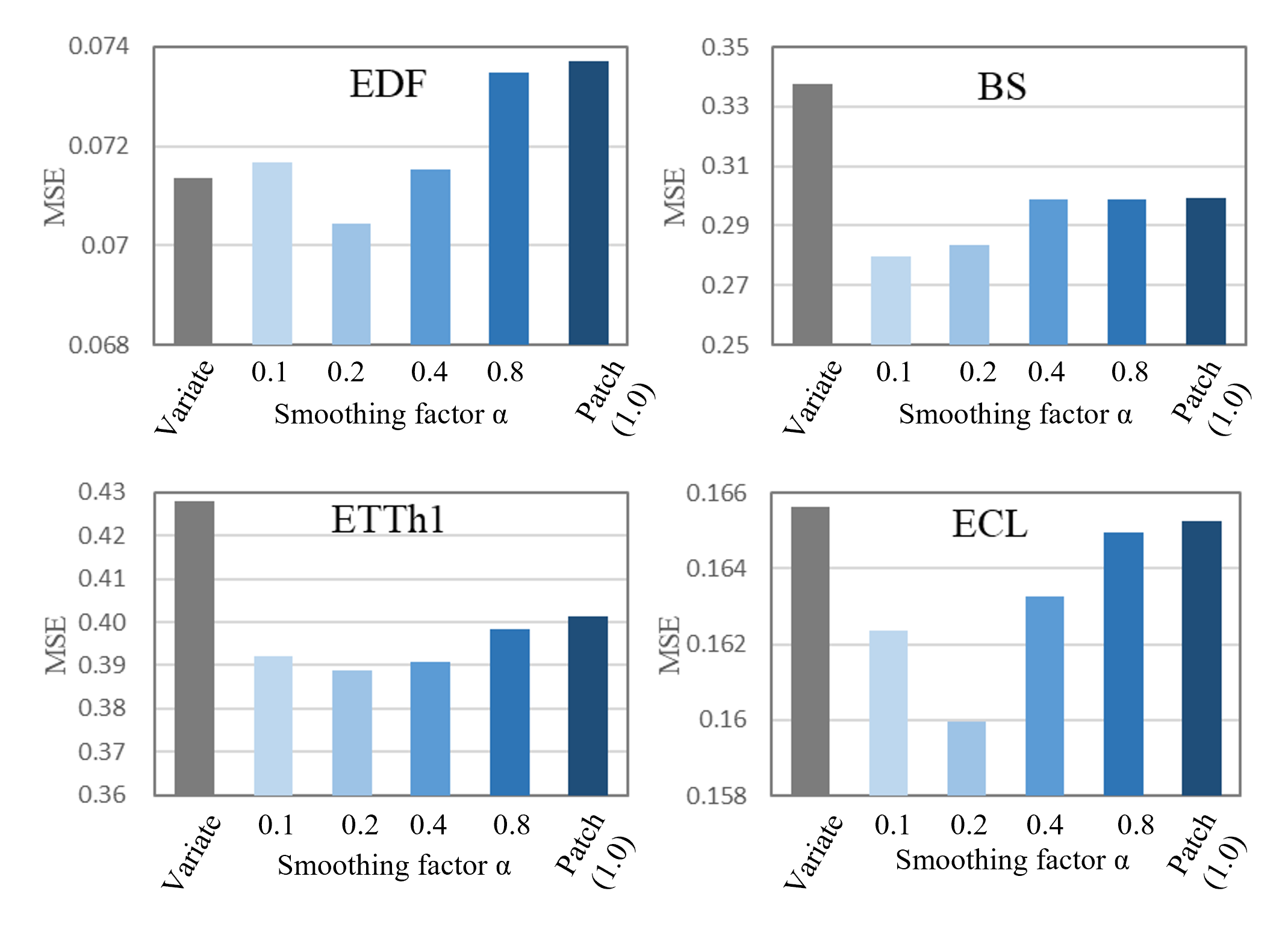}
    \caption{The performance at different levels of granularity in cross-variate attention. The performance at ``Variate'' level is represented by TimeXer. Strong Attention Score Smoothing (around $\alpha=0.2$) outperforms both variate-level and patch-level approaches. The performance is evaluated with a fixed random seed.}
    \label{fig:abl_ass}
\end{figure}

Attention Score Smoothing captures locally accurate patch-level cross-variate dependencies and refines them into global variate-level dependencies. Note that a large smoothing factor $\alpha$ emphasizes patch-level dependencies (with $\alpha=1.0$ corresponding to no smoothing), while a smaller $\alpha$ emphasizes variate-level dependencies through heavier smoothing. Figure~\ref{fig:abl_ass} illustrates the performance at different levels of granularity in cross-variate attention. The performance of the variate-level approach is represented by TimeXer. We observe that the suitability of variate-level versus patch-level approaches is dependent on the dataset. However, strong Attention Score Smoothing (around $\alpha=0.2$) strikes a balance between the two, highlighting the versatile effectiveness of this mechanism in both covariate-informed and multivariate forecasting.

\subsubsection{Representation Analysis}
\begin{figure}[h]
    \includegraphics[trim=7 8 5 10, clip, width=0.95\linewidth]
    {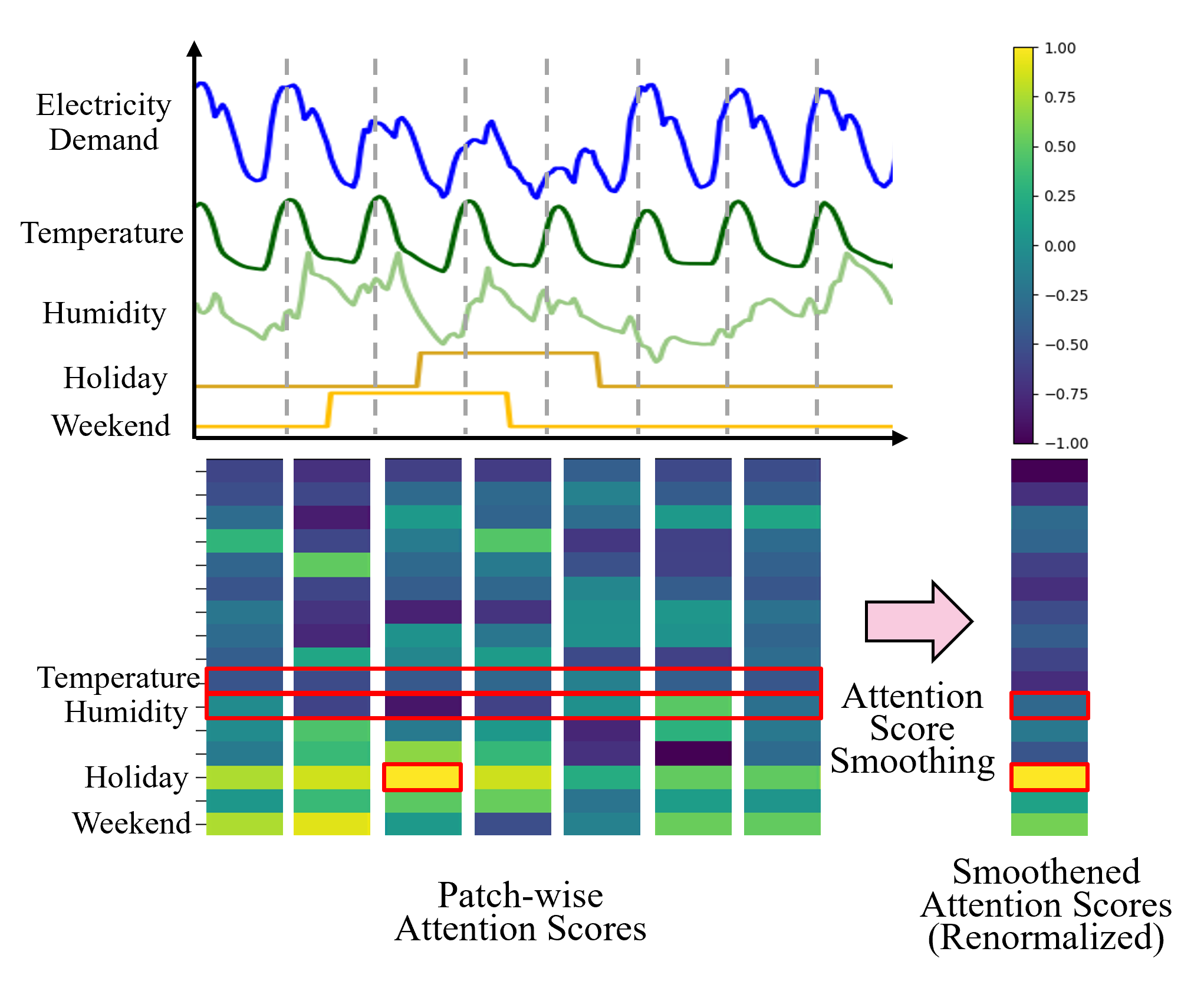}
    \caption{Visualization of attention scores between the target ``Electricity Demand'' and other variables observed in the EDF dataset. By applying Attention Score Smoothing, the smoothed scores capture a strong dependency on ``Holiday'' while mitigating noisy interactions caused by fluctuation in ``Humidity''. }
    \label{fig:representation}
\end{figure}

To uncover the covariate usage mechanism of CITRAS further, we present the visualization of the cross-variate attention scores between the target variable and other variables in the EDF dataset in Figure~\ref{fig:representation}.
It is observed that patch-wise attention scores effectively capture the local negative impact of ``Holiday'' on the target variable (``Electricity Demand'').
However, these scores are also susceptible to local disturbances, as demonstrated by the relatively fluctuating scores of ``Humidity'' compared to the ``Temperature'' that shows stable correlation with the target variable.
By applying Attention Score Smoothing, the obtained scores effectively convey the strong dependency on ``Holiday'' from past steps and mitigate the noisy interaction caused by the fluctuating scores of ``Humidity''.
In this way, CITRAS enjoys the advantages of both patch-level and variate-level approaches while offering improved interpretability.

\subsubsection{Computational Efficiency}
\label{sec:efficiency}
We analyze the computational complexity of CITRAS in comparison to other models.
Consider the covariate-informed setting where the number of target variables is $1$, the number of covariates (including observed and known covariates) is $C$, and the number of patches of the target variable is $N$.
CITRAS calculates cross-variate attention between the target and covariates at each patch step, resulting in a computational complexity of $O(CN)$.
Additionally, it applies cross-time attention to each variate separately, incurring a complexity of $O(CN^2)$.
This is favorable compared to the Any-variate Attention adopted by Timer-XL, which incurs $O(C^2N^2)$ to flexibly accommodate covariates.
Furthermore, the $O(CN)$ complexity of cross-variate attention is advantageous when the number of covariates $C$ increases, as the multivariate models like iTransformer that do not discriminate between target and covariate incur the cost of $O(C^2)$.

\begin{figure}[h]
    \centering
    \includegraphics[trim=10 7 5 5, clip, width=0.8\linewidth]{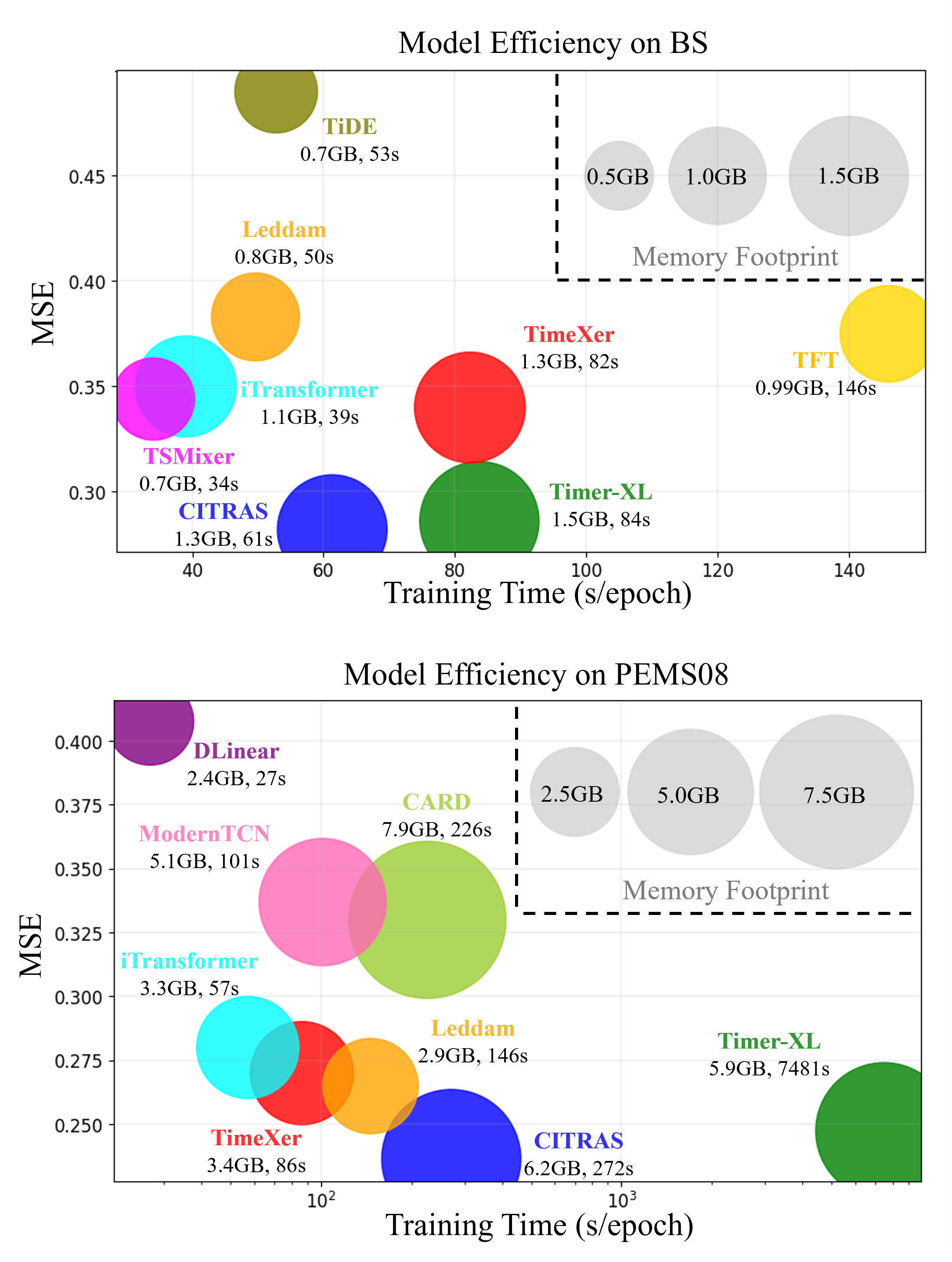}
    \caption{Comparison of model efficiency on the BS and PEMS08 datasets. Note that the x-axis is in logarithmic scale for the PEMS08 result.}
    \label{fig:efficiency}
\end{figure}

Besides the theoretical analysis, we demonstrate the empirical model efficiency on the BS dataset and PEMS08 dataset in Figure~\ref{fig:efficiency}.
We use the best model configuration for each model and evaluate using the same batch size for a fair comparison.
The results in the BS dataset underscore the efficient model design of CITRAS.
In the PEMS08 dataset, we observe the trade-off between training time and model performance in terms of MSE. 
Among them, CITRAS achieves the best performance without significant compromise in efficiency.

\section{Conclusion}
Considering the critical roles of covariates in practical time series forecasting, we proposed CITRAS, a decoder-only Transformer that flexibly leverages multiple target variables, observed covariates, and known covariates within a unified framework. To this end, we introduced two novel mechanisms in patch-wise cross-variate attention: Key-Value Shift, which incorporates future known covariate information into autoregressive forecasting, and Attention Score Smoothing, which refines locally captured patch-level dependencies into more stable variate-level relationships. Through extensive experiments, CITRAS demonstrated strong and consistent performance in both covariate-informed and multivariate forecasting settings, ranking first more frequently than other state-of-the-art models and highlighting its ability to effectively model cross-variate and cross-time dependencies.

Despite these promising results, several limitations remain. First, CITRAS operates in a supervised setting and requires dataset-specific training to achieve optimal performance, which may limit its applicability in low-data regimes or unseen domains. Second, in its current form, CITRAS does not explicitly distinguish between continuous and categorical covariates. While this design choice simplifies the architecture, it may restrict the expressive capacity when modeling discrete or event-driven covariates, such as calendar indicators.

Future work will address these limitations by exploring large-scale pre-training strategies to improve generalization across datasets and tasks, as well as by incorporating categorical-aware encoding schemes, such as target encoding, to better capture the semantics of categorical covariates within the decoder-only Transformer framework.
\section{Appendix}

\subsection{Standard Deviation of Covariate-Informed Forecasting}\label{sec:covariate-std}
Table~\ref{tab:covariate-result-std} reports the standard deviation of MSE and MAE across three random seeds for all models in the covariate-informed forecasting experiments, corresponding to the results presented in Table~\ref{tab:covariate-result} in Section~\ref{sec:covariate-forecasting}.
The results indicate that CITRAS exhibits stable performance across different random initializations.

\begin{table*}[ht]
\begin{threeparttable}
\caption{Standard deviation of the covariate-informed forecasting results across three random seeds.}
\label{tab:covariate-result-std}
\setlength{\tabcolsep}{3.5pt}
\centering
{\fontsize{9pt}{10.5pt}\selectfont
\begin{tabular}
{p{10pt}|c|cccccccccccccccc} 

\toprule[1.2pt]
\multicolumn{2}{c|}{{Model}}   &
\multicolumn{2}{c}{{\textbf{CITRAS}}} &
\multicolumn{2}{c}{{TFT}} &
\multicolumn{2}{c}{{TSMixer-Ext}} &
\multicolumn{2}{c}{{TimeXer}} &
\multicolumn{2}{c}{{TiDE}} &
\multicolumn{2}{c}{{Timer-XL}} &
\multicolumn{2}{c}{{iTrans.}} &
\multicolumn{2}{c}{{Leddam}}\\

\cmidrule(lr){3-4}\cmidrule(lr){5-6}\cmidrule(lr){7-8}\cmidrule(lr){9-10}
\cmidrule(lr){11-12}\cmidrule(lr){13-14}\cmidrule(lr){15-16}\cmidrule(lr){17-18}
\multicolumn{2}{c|}{{Metric}} &
{MSE} & {MAE} &
{MSE} & {MAE} &
{MSE} & {MAE} &
{MSE} & {MAE} &
{MSE} & {MAE} &
{MSE} & {MAE} &
{MSE} & {MAE} &
{MSE} & {MAE}\\ 

\toprule
\multirow{7}{*}{\vspace{-28pt} \rotatebox{90}{\scalebox{1.4}{\hspace{15pt}w/ known}}}
& {EPF--NP}
& {0.001} & {0.002}
& {0.004} & {0.005}
& {0.005} & {0.003}
& {0.001} & {0.001}
& {0.001} & {0.001}
& {0.005} & {0.003}
& {0.005} & {0.003}
& {0.025} & {0.017} \\

& {EPF-PJM}
& {0.001} & {0.000}
& {0.004} & {0.004}
& {0.000} & {0.002}
& {0.001} & {0.002}
& {0.000} & {0.001}
& {0.001} & {0.001}
& {0.001} & {0.000}
& {0.004} & {0.002} \\

& {EPF--BE}
& {0.009} & {0.005}
& {0.031} & {0.004}
& {0.048} & {0.010}
& {0.003} & {0.001}
& {0.003} & {0.001}
& {0.035} & {0.014}
& {0.009} & {0.001}
& {0.003} & {0.002} \\

& {EPF--FR}
& {0.004} & {0.001}
& {0.008} & {0.005}
& {0.026} & {0.004}
& {0.002} & {0.002}
& {0.004} & {0.002}
& {0.022} & {0.003}
& {0.006} & {0.001}
& {0.002} & {0.001} \\

& {EPF--DE}
& {0.002} & {0.001}
& {0.007} & {0.004}
& {0.002} & {0.001}
& {0.014} & {0.002}
& {0.008} & {0.005}
& {0.001} & {0.002}
& {0.008} & {0.005}
& {0.013} & {0.004} \\

& {EDF}
& {0.001} & {0.001}
& {0.002} & {0.003}
& {0.004} & {0.007}
& {0.001} & {0.001}
& {0.000} & {0.001}
& {0.002} & {0.002}
& {0.001} & {0.001}
& {0.001} & {0.002} \\

& {BS}
& {0.005} & {0.003}
& {0.068} & {0.037}
& {0.014} & {0.011}
& {0.002} & {0.001}
& {0.002} & {0.002}
& {0.001} & {0.001}
& {0.009} & {0.005}
& {0.007} & {0.004} \\

\toprule[1.2pt]
\multicolumn{2}{c|}{{Model}}   &
\multicolumn{2}{c}{{\textbf{CITRAS}}} &
\multicolumn{2}{c}{{CARD}} &
\multicolumn{2}{c}{{ModernTCN}} &
\multicolumn{2}{c}{{TimesNet}} &
\multicolumn{2}{c}{{Cross.}} &
\multicolumn{2}{c}{{FITS}} &
\multicolumn{2}{c}{{DLinear}} &
\multicolumn{2}{c}{{PatchTST}}\\

\cmidrule(lr){3-4}\cmidrule(lr){5-6}\cmidrule(lr){7-8}\cmidrule(lr){9-10}\cmidrule(lr){11-12}\cmidrule(lr){13-14}\cmidrule(lr){15-16}\cmidrule(lr){17-18}
\multicolumn{2}{c|}{{Metric}} & {MSE} & {MAE} & {MSE}    & {MAE}   & {MSE}    & {MAE}    & {MSE}    & {MAE} & {MSE}    & {MAE} & {MSE}    & {MAE} & {MSE}    & {MAE} & {MSE}    & {MAE}\\ 

\toprule
\multirow{7}{*}{\vspace{-28pt} \rotatebox{90}{\scalebox{1.4}{\hspace{13pt}w/o known}}} & {EPF--NP} & {0.001} & {0.001} & {0.003} & {0.003} & {0.002} & {0.001} & {0.013} & {0.003} & {0.002} & {0.004} & {0.000} & {0.000} & {0.000} & {0.000} & {0.004} & {0.003} \\

& {EPF-PJM} & {0.002} & {0.001} & {0.004} & {0.004} & {0.001} & {0.000} & {0.004} & {0.003} & {0.004} & {0.002} & {0.000} & {0.000} & {0.000} & {0.000} & {0.004} & {0.004} \\

& {EPF--BE} & {0.001} & {0.001} & {0.011} & {0.005} & {0.002} & {0.001} & {0.011} & {0.005} & {0.013} & {0.013} & {0.000} & {0.000} & {0.001} & {0.001} & {0.003} & {0.003} \\

& {EPF--FR} & {0.001} & {0.001} & {0.001} & {0.002} & {0.002} & {0.001} & {0.012} & {0.006} & {0.026} & {0.003} & {0.001} & {0.001} & {0.000} & {0.000} & {0.006} & {0.003} \\

& {EPF--DE} & {0.011} & {0.004} & {0.010} & {0.004} & {0.001} & {0.001} & {0.049} & {0.014} & {0.034} & {0.001} & {0.000} & {0.000} & {0.004} & {0.002} & {0.007} & {0.005} \\

& {EDF} & {0.001} & {0.001} & {0.002} & {0.003} & {0.001} & {0.001} & {0.008} & {0.009} & {0.001} & {0.002} & {0.000} & {0.000} & {0.000} & {0.000} & {0.001} & {0.000} \\

& {BS} & {0.000} & {0.001} & {0.004} & {0.002} & {0.002} & {0.001} & {0.014} & {0.004} & {0.010} & {0.006} & {0.000} & {0.001} & {0.000} & {0.000} & {0.003} & {0.003} \\

\bottomrule[1.2pt]

\end{tabular}}
\begin{tablenotes}[para,flushleft]
    Standard deviation is computed across three random seeds. The experimental settings are identical to those in Table~\ref{tab:covariate-result}.
\end{tablenotes}
\end{threeparttable}

\end{table*}

\subsection{Validation Curves of Covariate-Informed Forecasting}
\begin{figure*}[ht]
\centering
\includegraphics[width=\linewidth]{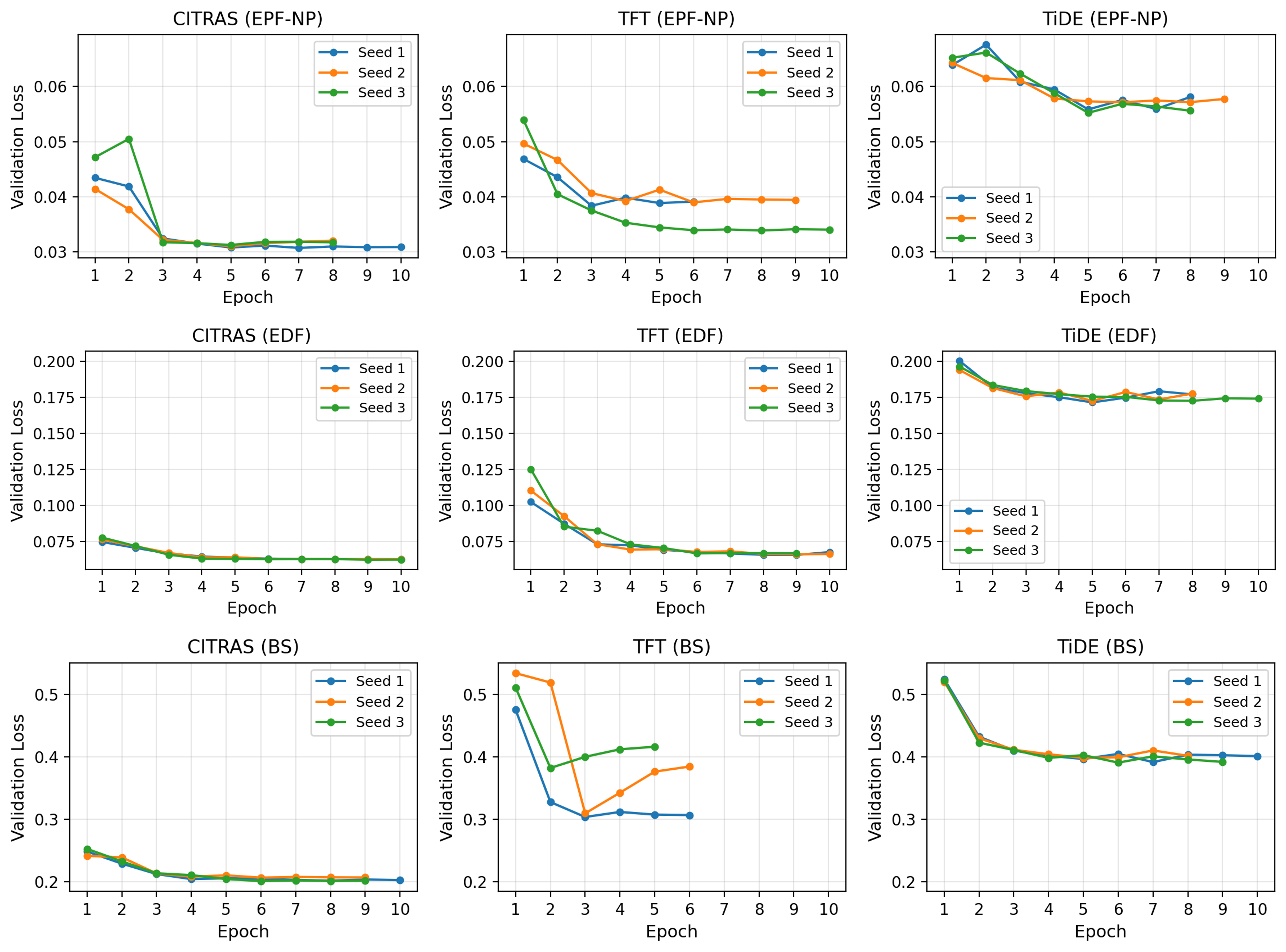}
\caption{Validation loss curves on EPF-NP (top), EDF (middle), and BS (bottom) for CITRAS (left), TFT (middle), and TiDE (right) across three random seeds.}
\label{fig:val-curve}
\end{figure*}

In our experiments, all models are trained with the same maximum training budget of 10 epochs, using early stopping based on validation performance. To justify this training budget, we report the validation MSE loss curves for representative covariate-informed datasets (EPF-NP, EDF, and BS) for CITRAS (Transformer-based), TFT (Transformer- and RNN-based), and TiDE (MLP-based) over three random seeds. The results in Figure~\ref{fig:val-curve} show that the validation loss typically stabilizes within the chosen budget (often earlier due to early stopping) regardless of the base architecture, supporting the choice of a maximum training budget of 10 epochs.

\subsection{Extra Results of Covariate-Informed Forecasting}
To further evaluate CITRAS against recent models, we compare it with Sonnet~\cite{shu2025sonnet}, which was presented at AAAI 2026, on covariate-informed forecasting tasks.
Sonnet is designed for settings with a univariate target and observed covariates only, and therefore cannot be applied to datasets with multiple target variables (e.g., the BS dataset) or to settings that involve known covariates.
For this reason, the comparison is limited to the EPF datasets and the EDF dataset under the ``w/o known'' setting.
The results in Table~\ref{tab:sonnet-comparison} show that CITRAS achieves competitive performance compared to this recent state-of-the-art model.
Notably, unlike Sonnet, CITRAS can additionally leverage future known covariates when they are available to further improve forecasting accuracy.
\begin{table}[ht]
\begin{threeparttable}
\caption{Comparison of CITRAS and Sonnet on covariate-informed forecasting without known covariates. Average MSE and MAE across three random seeds are reported.}
\label{tab:sonnet-comparison}
\setlength{\tabcolsep}{3.5pt}
\centering
{\fontsize{9pt}{10.5pt}\selectfont
\begin{tabular*}{0.8\linewidth}{@{\extracolsep{\fill}}p{15pt}|c|cccc}

\toprule[1.2pt]
\multicolumn{2}{c|}{{Model}}   &
\multicolumn{2}{c}{{\textbf{CITRAS}}} &
\multicolumn{2}{c}{{Sonnet}}\\

\cmidrule(lr){3-4}\cmidrule(lr){5-6}
\multicolumn{2}{c|}{{Metric}} &
{MSE} & {MAE} &
{MSE} & {MAE}\\

\toprule
\multirow{6}{*}{\vspace{-10pt} \rotatebox{90}{\scalebox{1.4}{\hspace{5pt}w/o known}}}
& {EPF--NP}
& \best{{0.227}} & \best{{0.260}}
& {0.250} & {0.281} \\

& {EPF--PJM}
& \best{{0.090}} & \best{{0.182}}
& {0.106} & {0.205} \\

& {EPF--BE}
& {0.405} & {0.276}
& \best{{0.394}} & \best{{0.263}} \\

& {EPF--FR}
& {0.407} & {0.219}
& \best{{0.400}} & \best{{0.218}} \\

& {EPF--DE}
& \best{{0.420}} & \best{{0.404}}
& {0.449} & {0.424} \\

& {EDF}
& {0.087} & \best{{0.206}}
& \best{{0.086}} & \best{{0.206}} \\

\bottomrule[1.2pt]

\end{tabular*}}
\begin{tablenotes}[para,flushleft]
    The input length and forecasting length are set to 168 and 24, respectively.
\end{tablenotes}
\end{threeparttable}
\end{table}

\subsection{Hyperparameter Sensitivity Analysis}
In addition to the sensitivity analysis on the smoothing factor $\alpha$ of Attention Score Smoothing (Section~\ref{sec:ass_param}), we further investigate the sensitivity of CITRAS to the patch size, which is a key hyperparameter in patch-based Transformer architectures.

Specifically, we evaluate CITRAS on the EPF-NP, EDF, and BS datasets using patch sizes of 12, 24, and 42, while fixing the input length and forecasting horizon to 168 and 24, respectively, following the same configuration as in Table~\ref{tab:covariate-result}. In CITRAS, the output length is aligned with the patch size through the KV Shift mechanism. As a result, when the patch size is shorter than the forecasting horizon (e.g., patch size = 12), recursive forecasting is applied to reach the full horizon, whereas for larger patch sizes (e.g., patch size = 42), predictions up to 24 steps are used for evaluation. 

\begin{figure*}[ht]
\centering
\begin{minipage}{0.32\textwidth}
    \centering
    \includegraphics[width=\linewidth]{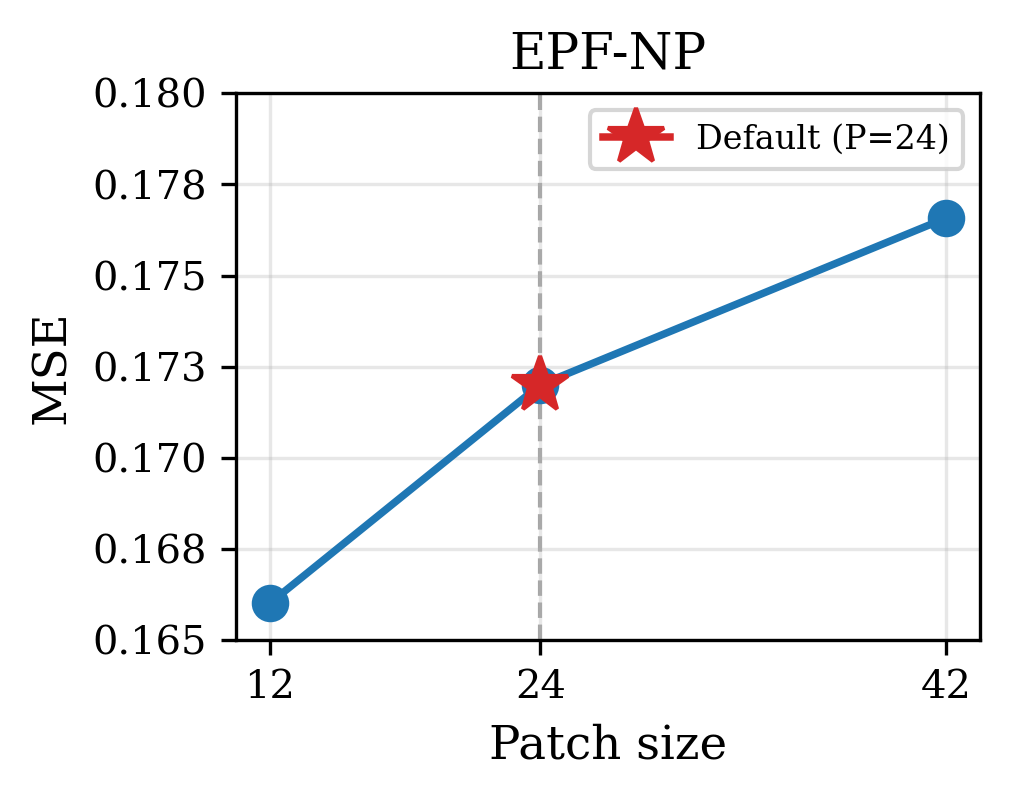}
\end{minipage}
\hfill
\begin{minipage}{0.32\textwidth}
    \centering
    \includegraphics[width=\linewidth]{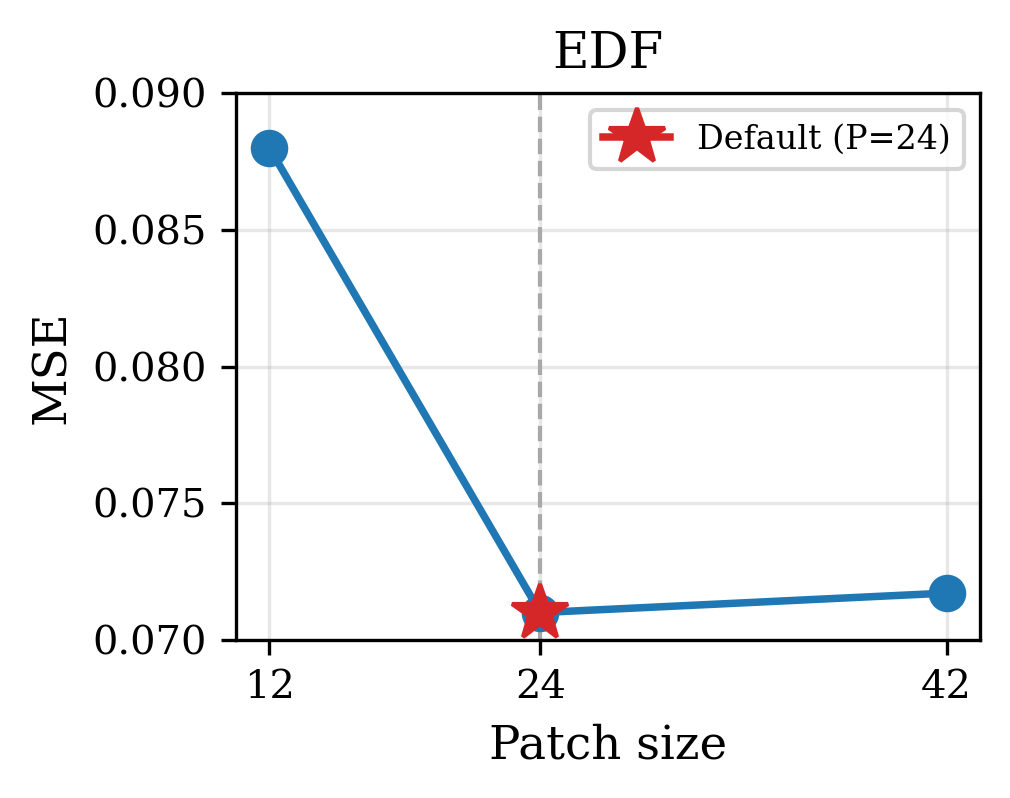}
\end{minipage}
\hfill
\begin{minipage}{0.32\textwidth}
    \centering
    \includegraphics[width=\linewidth]{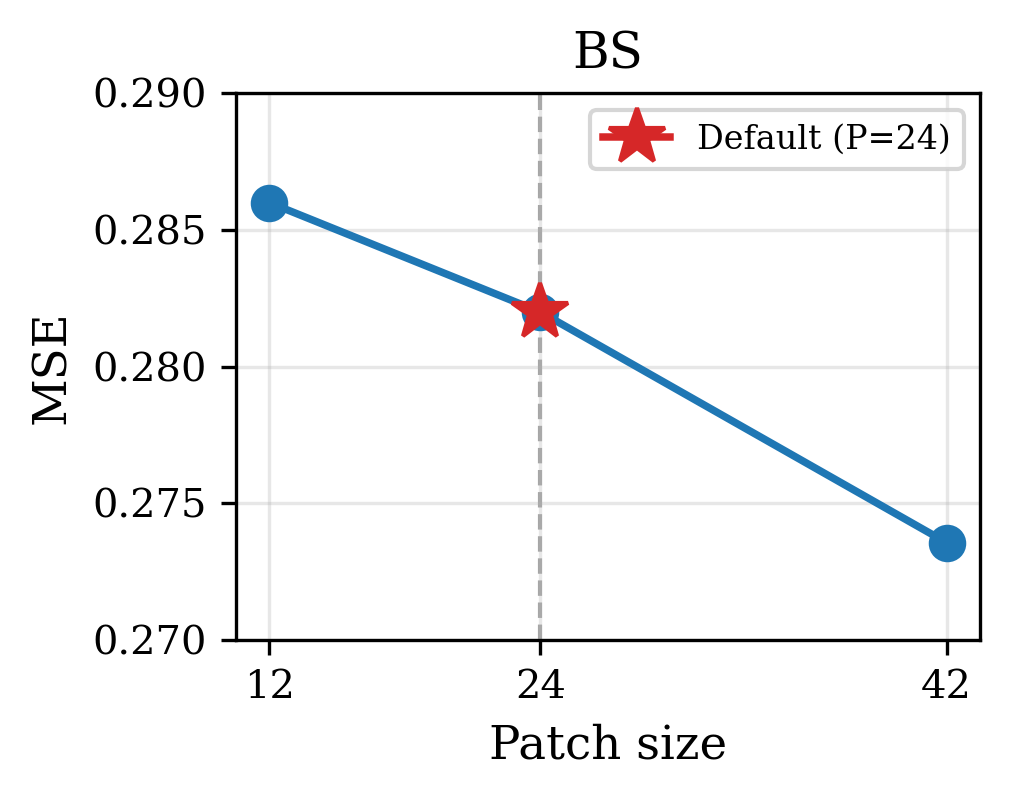}
\end{minipage}
\caption{Patch size sensitivity analysis on the EPF-NP, EDF, and BS datasets. The dashed line indicates the default patch size of 24 adopted in all experiments.}
\label{fig:patch_sensitivity}
\end{figure*}

As shown in Figure~\ref{fig:patch_sensitivity}, CITRAS is generally not overly sensitive to the choice of patch size within a reasonable range. One exception is the EDF dataset under a small patch size of 12, which exhibits relatively worse performance compared to larger patch sizes. This behavior is expected for electricity demand forecasting, which exhibits strong daily seasonality and therefore benefits from a patch size that spans a full daily cycle. Overall, the adopted patch size of 24 yields stable and competitive performance across all datasets, supporting the reasonableness of this choice under a fixed hyperparameter setting.

\subsection{Model Extension Details}\label{sec:extension}
In this section, we describe the implementation details of the extensions for the baseline models (Timer-XL, iTransformer, and Leddam) introduced in Section~\ref{sec:covariate-baselines}, as well as the late fusion variant of CITRAS analyzed in the ablation study in Section~\ref{sec:ablation}.
\paragraph{Extension of Timer-XL.}
Timer-XL~\cite{liu2025timerxl} originally supports only observed covariates alongside multivariate targets.
To enable the use of known covariates, we modify the causal masking scheme in its TimeAttention module, while keeping all other components identical to the original implementation.

In TimeAttention, time series variables—including the target and covariates—are tokenized into patch-level tokens and flattened into a single sequence, allowing self-attention to jointly model dependencies across time steps and variables.
In the original masking scheme, a target token at patch step $i$ is permitted to attend to tokens at the same or earlier patch steps ($\leq i$) from all variables, while all future tokens are strictly masked.

To incorporate known covariates, we relax this masking scheme by allowing target token embeddings at patch step $i$ to attend to the one-step-ahead future tokens ($i+1$) corresponding to known covariates.
Future tokens of the target variable and observed covariates remain masked.
This design allows Timer-XL to exploit future covariate information that is available at prediction time, without violating the temporal causality of the target series.

\paragraph{Extensions of iTransformer and Leddam.}
iTransformer~\cite{liu2023itransformer} and Leddam~\cite{yu2024leddam} are originally designed as multivariate forecasting models, where each variable is represented by embedding its entire historical sequence into a single token and modeling inter-variable dependencies through attention mechanisms.
In their original formulations, the target variable with length $T$ is embedded into a $D$-dimensional representation using a variate embedding layer with weights of size $T \times D$.

To incorporate known covariates whose available sequence length extends beyond the target horizon, we introduce an additional embedding layer specifically for known covariates.
This embedding layer maps the extended covariate sequences of length $T+S$ into the same $D$-dimensional space using weights of size $(T+S) \times D$, allowing the model to ingest future covariate information.
Afterwards, these embeddings are treated in the same manner as the original variable embeddings and processed by the model following the original implementation.

\paragraph{Late Fusion Variant of CITRAS.}
As discussed in the ablation study in Section~\ref{sec:ablation}, we evaluate a simpler alternative to KV Shift, referred to as the ``w/ Late fusion'' variant, which incorporates future known covariates without modifying the cross-variate attention mechanism. In this variant, the embedding layers are identical to those used in the full CITRAS model, producing target, observed-covariate, and known-covariate token embeddings, respectively. Cross-time attention with causal masking and cross-variate attention are then applied in the same manner as in the full model; however, KV Shift is not employed, and the target token is therefore unable to attend to future known-covariate tokens within the attention layers.

Instead, future known covariates are incorporated through an additional learnable fusion layer placed immediately before the final projection layer. Specifically, the last target token $\mathbf{H}_{N_{tgt}}^{tgt,c}\in\mathbb{R}^{1\times D}$ and the one-step-ahead known-covariate tokens $\mathbf{H}_{N_{tgt}+1}^{knw,:}\in\mathbb{R}^{C_{knw}\times D}$ are used as inputs. The known-covariate tokens are first pooled along the covariate dimension to obtain a single $\mathbb{R}^{1\times D}$ representation, concatenated with $\mathbf{H}_{N_{tgt}}^{tgt,c}$, and projected to $\mathbb{R}^{1\times D}$ through a linear fusion layer of size $2D\!\rightarrow\!D$. The fused target representation is then passed to the original projection layer to generate the final prediction. All components of this variant are trained from scratch.

\bibliographystyle{IEEEtran}
\bibliography{ref}

\EOD

\end{document}